\documentclass{article}

\usepackage[preprint]{neurips_2026}

\usepackage[utf8]{inputenc}
\usepackage[T1]{fontenc}
\usepackage{caption}
\usepackage{hyperref}
\usepackage{url}
\usepackage{booktabs}
\usepackage{amsfonts}
\usepackage{amsmath}
\usepackage{amssymb}
\usepackage{microtype}
\usepackage[most]{tcolorbox}
\usepackage{xcolor}
\usepackage{enumitem}
\usepackage{graphicx}
\usepackage{wrapfig}

\usepackage{pifont}
\usepackage{multirow}
\newcommand{\cmark}{\ding{51}}

\title{BenchCAD: A Comprehensive, Industry-Standard Benchmark for Programmatic CAD}

\author{
    \textbf{Haozhe Zhang$^{1,*,\dagger}$} \quad
    \textbf{Kaichen Liu$^{2,*}$} \quad
    \textbf{Miaomiao Chen$^{1,*}$} \\
    \textbf{Lei Li$^{1}$} \quad
    \textbf{Shaojie Yang$^{2}$} \quad
    \textbf{Cheng Peng$^{1}$} \quad
    \textbf{Hanjie Chen$^{3,\dagger}$} \\
    \\
    $^1$University of Virginia \\
    $^2$University of California, San Diego \\
    $^3$Rice University \\
    \\
    $^*$Equal contribution. \quad
    $^\dagger$Corresponding authors. \\
    \texttt{hz5sq@virginia.edu, hanjie@rice.edu}
}

\date{May 2026}
\begin{document}

\maketitle

\begin{abstract}
Industrial Computer-Aided Design (CAD) code generation requires models to produce executable parametric programs from visual or textual inputs. 
Beyond recognizing the outer shape of a part, this task involves understanding its 3D structure, inferring engineering parameters, and choosing CAD operations that reflect how the part would be designed and manufactured. Despite the promise of Multimodal large language models (MLLMs) for this task, they are rarely evaluated on whether these capabilities jointly hold in realistic industrial CAD settings. We present BenchCAD, a unified benchmark for industrial CAD reasoning. BenchCAD contains 17,900 execution-verified CadQuery programs across 106 industrial part families, including bevel gears, compression springs, twist drills, and other reusable engineering designs. It evaluates models through visual question answering, code question answering, image-to-code generation, and instruction-guided code editing, enabling fine-grained analysis across perception, parametric abstraction, and executable program synthesis. Across 10+ frontier models, BenchCAD shows that current systems often recover coarse outer geometry but fail to produce faithful parametric CAD programs. Common failures include missing fine 3D structure, misinterpreting industrial design parameters, and replacing essential operations such as sweeps, lofts, and twist-extrudes with simpler sketch-and-extrude patterns. Fine-tuning and reinforcement learning improve in-distribution performance, but generalization to unseen part families remains limited. These results position BenchCAD as a benchmark for measuring and improving the industrial readiness of multimodal CAD automation.
Released under CC-BY-4.0. \\
Project page:
\url{https://benchcad.github.io/BenchCAD_webpage/}.
\end{abstract}
% \lei{Abstract is a bit long (taking four sentences before reaching BenchCAD; three coupled capabilities, four tasks, three failure modes - a bit overwhelming in abstract), and should try to fit into one paragraph.}

\section{Introduction}
\label{sec:intro}

\begin{figure*}[t]
  \centering
  \includegraphics[width=\linewidth]{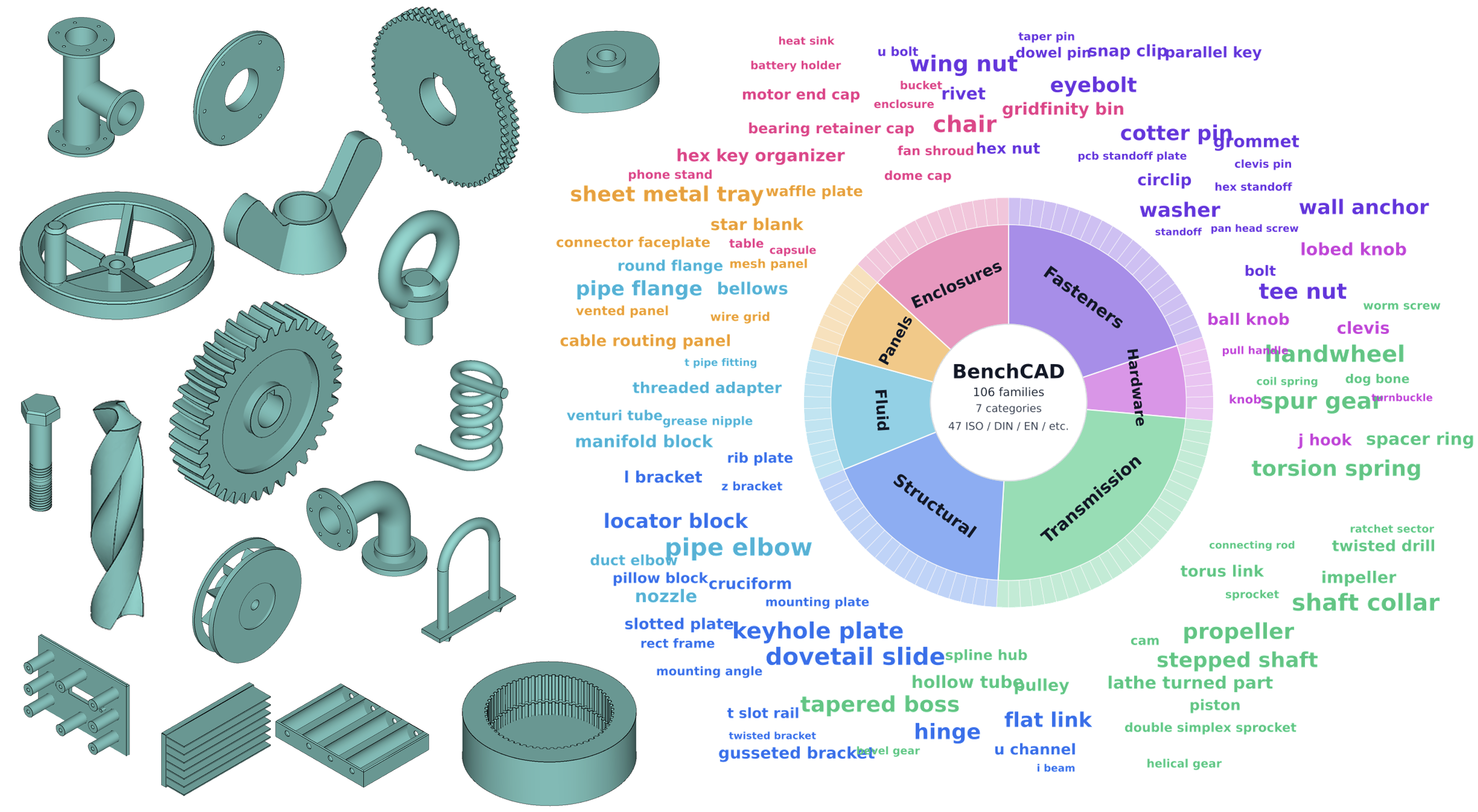}
  \caption{\textbf{BenchCAD overview.} BenchCAD is a unified, capability-decomposed evaluation framework for industrial CAD reasoning, consisting of 17{,}900 expert-verified parametric CadQuery parts (\emph{left}) drawn from 106 industrial families spanning fasteners, transmission components, structural elements, fluid fittings, panels, hardware, and enclosures. The 7-category functional taxonomy (\emph{right}) covers 49\% of families anchored to ISO/DIN/EN/ASME/IEC standards, with each part realised as an executable parametric program testable for geometry, parameters, and edits.
  % \hc{make the left part smaller; enlarge the font in the right part.}
  % \lei{Make the caption self-contained, i.e., a small paragraph. Could be something like BenchCAD is a unified evaluation framework for large-model CAD capability, consisting of XXX expert-verified parts (Left) across 7-category... (Right)}
  }
  \label{fig:teaser}
  
  \vspace{-1.5em}
\end{figure*}

% P1: MLLMs general capability + CAD
Multimodal large language models (MLLMs)~\citep{OpenAI2026GPT53Instant,Anthropic2026Claude47,Google2026Gemini31} now combine visual perception, code generation, and multi-step reasoning in a single interface, performing tasks that previously required dedicated systems.
A natural next question is whether such models can move beyond recognizing visual content to producing executable programs that define and modify physical objects.
Computer-aided design (CAD) provides a canonical testbed for this question.
Unlike a mesh, point cloud, or rendered image, a CAD model is typically an editable parametric program: geometric operations such as extrusion, cutting, sweeping, filleting, and patterning construct a solid object through variables and constraints~\citep{CadQuerySoftware}.
A capable CAD agent must therefore connect visual evidence, 3D structure, symbolic operations, and executable program synthesis.

% P2: How prior assesses these capabilities
Real industrial designs are rarely arbitrary shapes.
They belong to reusable families such as gears, springs, brackets, fasteners, pipes, and bearings (Fig.~\ref{fig:teaser}), where small local features and parameter relations decide whether a part is merely visually similar or actually useful as an editable design.
A spring is not fully specified by its helical silhouette: its end coils, pitch profile, and cross-section encode design intent.
Similarly, a gear-like outline is insufficient without the correct tooth construction and parameterization.
\textbf{The geometry passes the eye but fails the caliper.}

% P3: How prior assesses these capabilities
Existing evaluations capture only parts of this problem.
Prior CAD code-generation benchmarks~\citep{Wu2021,Khan2024,Guan2025,Rukhovich2025,Kolodiazhnyi2026,Elistratov2026} score whether a model's rendered output matches a target shape, typically through a single end-to-end geometric metric such as IoU or Chamfer distance, while program editing~\citep{Alrashedy2025,CADEditor2025} and design question answering are studied in isolation.
However, two programs may produce roughly similar outer envelopes while differing substantially in editability, operation choice, and engineering detail, so a single shape-matching score can overestimate capability and obscure which sub-ability --- visual perception, parametric abstraction, or code synthesis --- drives the remaining gap.

% P4: Research problem + Motivation
This leaves open how to evaluate CAD agents beyond rendered shape fidelity.
For practical use, the relevant question is not only ``does the rendered geometry match?'', but also ``does the model understand the operations, parameters, constraints, and editable structure that produced it?''
We therefore formalize CAD reasoning as a four-level capability hierarchy, from part-level visual recognition to CAD-operation understanding, parametric abstraction, and executable code synthesis (Fig.~\ref{fig:capability_hierarchy}).
This decomposition is important because CAD failures are often compositional: a model may identify the family but pick the wrong operation, infer the rough scale but miss a standard relation, or edit the requested feature while changing unrelated parts.

% P5: Introduce BenchCAD
We introduce BenchCAD, a capability-decomposed benchmark for executable, editable, and constraint-aware CAD reasoning.
BenchCAD contains 17{,}900 execution-verified CadQuery programs across 106 industrial part families covering twist drills, bevel gears, compression springs, propellers, brackets, flanges, eye bolts, fasteners, and other reusable industrial designs (Fig.~\ref{fig:teaser}).
It exercises a substantially broader CadQuery operation surface than prior released corpora, including helical sweeps, lofts, twist-extrudes, and parametric involute-gear construction.
Rather than sampling unconstrained primitive shapes, BenchCAD instantiates structured part families with meaningful parameter relations and standard-derived dimensions (Fig.~\ref{fig:dataset_overview}).
It evaluates three practical CAD-agent workflows through four task families: \textsc{Vision2Code} (\texttt{img2cq}, image-to-code generation from multi-view renders), \textsc{Vision QA} and \textsc{Code QA} (\texttt{qa\_img}, \texttt{qa\_code}, design question answering from visual or program inputs), and \textsc{Code Edit} (\texttt{edit\_code}, instruction-guided program editing).

% P7: Key insights
Across 10+ frontier MLLMs and open CAD-specialized baselines~\citep{Kolodiazhnyi2026,Elistratov2026}, BenchCAD reveals a consistent gap between apparent geometric similarity and true parametric understanding: current models often recognize the global part family but miss local engineering details, choose simplified or incorrect CAD operations, or fail to apply localized edits faithfully.
The matched comparison between image- and code-based QA further shows that visual recognition alone does not imply reliable parametric abstraction.
Supervised fine-tuning (SFT) and reinforcement learning (RL) on BenchCAD improve operation coverage and executable generation, but substantial out-of-distribution gaps remain, especially for held-out industrial families requiring advanced operations and precise design constraints.

\paragraph{Contributions.}
(i) We release BenchCAD, a domain-expert-verified benchmark of 17{,}900 executable CadQuery programs across 106 industrial part families with multi-view renders, design QA pairs, and curated edit examples.
(ii) We propose a unified CAD-agent evaluation framework covering image-to-code generation, image- and code-based QA, and instruction-guided code editing.
(iii) We provide a capability-decomposed evaluation of frontier MLLMs and open CAD-specialized models, showing that current systems remain limited by local detail recognition, CAD operation reasoning, parametric abstraction, and edit fidelity.
All data are released under \texttt{BenchCAD/BenchCAD} on Hugging Face under the CC-BY-4.0 license, and the evaluation code is released under \texttt{BenchCAD/BenchCAD-main} on GitHub under the MIT license.

\section{Related Work}
\label{sec:related}

\begin{wraptable}{r}{0.42\linewidth}
  \vspace{-\baselineskip}
  \centering
  \caption{\textbf{BenchCAD positioning.} Benchmark overview for MLLM CAD capabilities.}
  \label{tab:advantages}
  % Inner tabular only — caption/label emitted by host wraptable in sec/related.tex
\footnotesize
\setlength{\tabcolsep}{4pt}
\renewcommand{\arraystretch}{1.05}
\begin{tabular}{@{}lc@{}}
\toprule
                                              & \textbf{BenchCAD} \\
\midrule
Primary purpose                               & evaluation        \\
                                              & benchmark         \\
Industry domain experts checked               & Yes               \\
Named industrial families                     & \textbf{106}      \\
ISO/DIN/EN/ASME/IEC codes                     & \textbf{47}       \\
Verified edit pairs                           & \textbf{748}      \\
Paired numeric-QA items                       & \textbf{2{,}400}  \\
Capability-decomposed task suite              & \textbf{\cmark}   \\
Rotation + scale-invariant scoring            & \textbf{\cmark}   \\
\bottomrule
\end{tabular}

  \vspace{-0.5\baselineskip}
\end{wraptable}

\paragraph{CAD code-generation models.}
Recent work establishes a common SFT$+$RL post-training pipeline on procedural sketch-and-extrude corpora.
DeepCAD~\citep{Wu2021} and Text2CAD~\citep{Khan2024} introduced command-sequence corpora; CAD-Coder~\citep{Guan2025} reformulated the data as CadQuery and added GRPO with a Chamfer reward; the CAD-Recode~\citep{Rukhovich2025}/cadrille~\citep{Kolodiazhnyi2026}/CADEvolve~\citep{Elistratov2026} lineage drove DeepCAD IoU to $\sim$92\% via multi-modal SFT and online RL on 1M--2.7M-script corpora.
Together they establish CadQuery code as a promising generation target for VLMs.
Two structural gaps remain: (i) training is dominated by sketch+extrude --- helical sweeps and parametric involute-gear construction are absent from every released corpus; and (ii) evaluation lacks family-level taxonomy and standard-table grounding, conflating visual perception, parametric abstraction, and code synthesis into a single end-to-end IoU score.

\paragraph{CAD evaluation benchmarks.}
The closest prior CAD code/edit benchmarks --- CADPrompt~\citep{Alrashedy2025}, CAD-Editor~\citep{CADEditor2025}, CADialogue~\citep{CADialogue2025}, and HistCAD~\citep{Dong2026} --- each cover a slice but none combines (i) execution-verified parametric edit pairs, (ii) a non-target preservation metric, and (iii) capability-decomposed sub-tasks. Detailed protocol comparison is in Appendix~\ref{app:related_extended}.

\paragraph{Position of BenchCAD.}
Rather than another training corpus, BenchCAD provides a unified, capability-decomposed evaluation framework for large-model CAD reasoning. It is the first public CAD benchmark to combine four properties simultaneously: (i) \emph{execution-verified at scale} (17{,}900 parts); (ii) \emph{standard-anchored} (49\% families bound to ISO/DIN/EN/ASME/IEC tables); (iii) \emph{operation-rich} (49 CadQuery operations including \texttt{helix}, \texttt{twistExtrude}, \texttt{polarArray}); (iv) \emph{capability-decomposed} (four tasks with image/code matched-contrast pairs that isolate visual recognition, parametric abstraction, and code synthesis, including the first verified parametric edit task). Table~\ref{tab:advantages} summarises the evaluation-side properties added; full per-axis comparison in Appendix~\ref{app:cmp_prior}.

% \hc{Figure 1 and 2 are not referred in the main context. I would also suggest moving Figure 2 to the Intro, and maybe switch the order of Figure 1 and 2.}
% \lei{Agree}

\section{BenchCAD: Dataset and Tasks}
\label{sec:dataset}

\begin{figure*}[t]
  \centering
  \includegraphics[width=\linewidth]{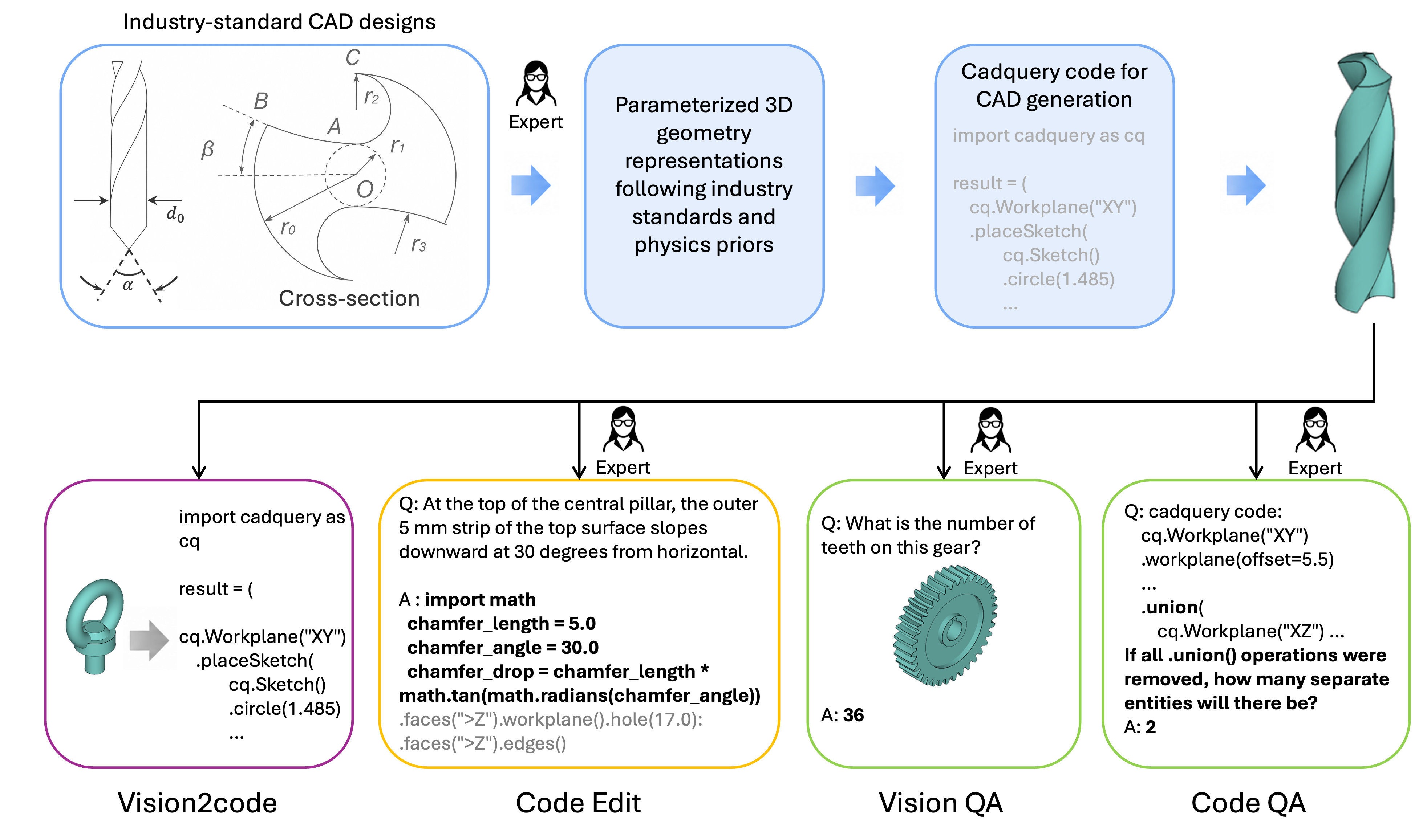}
  \caption{\textbf{BenchCAD generation pipeline and task suite.} \emph{Top:} parts originate from industry-standard engineering designs (e.g., DIN~338 twist-drill cross-section), are realised as parameterised 3D geometry that respects standard parameter relations and physical priors, and are emitted as executable CadQuery code with verified geometry. \emph{Bottom:} the four BenchCAD evaluation task categories operationalised on these parts --- \texttt{img2cq} (image-to-code), \texttt{edit\_code} (instruction-guided program editing), \texttt{qa\_img} (image-based design QA), and \texttt{qa\_code} (code-based design QA).}
  \label{fig:dataset_overview}
\vspace{-1.2em}
\end{figure*}

\subsection{Dataset}

\paragraph{Design principles.} BenchCAD rests on four principles.
\emph{(P1) Expert-generated geometry}: every family is hand-crafted by domain experts directly from industrial standards, who solve the standard-mandated geometric equations to produce parameterised CAD models that strictly respect engineering conventions and inter-parameter constraints.
\emph{(P2) Standard-table anchoring}: where a part has an industrial counterpart, parameters sample from real specification tables (e.g., ISO 22 V-belt cross-sections, DIN 338 twist-drill diameters, ISO 23509 bevel-gear pitch--module relations; full list in Appendix~\ref{app:standards}); 49\% of BenchCAD families (52/106) are standard-anchored, drawing from 47 unique ISO/DIN/EN/ASME/IEC codes.
% \lei{Are these exhaustive or just a few examples?}
\emph{(P3) Family-level taxonomy}: records are grouped into 106 named families (\texttt{coil\_spring}, \texttt{helical\_gear}, $\ldots$); a family may further branch into \emph{subfamilies} that capture mating or construction variants of the same part type (e.g., male/female fasteners). Each (sub)family is implemented by a small Python module exposing a typed parameter schema, sampler, validator, and deterministic builder --- making coverage measurable and per-family analysis trivial (subfamily definition in Appendix~\ref{app:datagen}).
\emph{(P4) Operation breadth}: BenchCAD exercises a substantially broader CadQuery operation surface than prior released corpora --- spanning primitives, 2D sketching, advanced solid ops (helical sweeps, lofts, shells), boolean composition, holes, arrays, and finishing features, including operations (\texttt{makeHelix}, \texttt{twistExtrude}, \texttt{polarArray}) rare or absent in DeepCAD/Fusion360-derived corpora (per-corpus breakdown in Appendix~\ref{app:ops}).

\paragraph{Generation pipeline.} Each family supports three difficulty tiers --- easy / medium / hard --- defined by parameter complexity, where higher tiers expand parameter ranges and activate optional features (full tier definition in Appendix~\ref{app:datagen}).
% \lei{How is difficulty defined? Seems not mentioned above.}
For each family$\times$subfamily$\times$tier bucket we sample parameters under standard-table constraints, emit the CadQuery program, and sandbox-execute it; programs that fail to compile, exceed a 30\,s runtime budget, or produce degenerate (zero or inverted) volume are quarantined (full failure-mode taxonomy in Appendix~\ref{app:datagen}).
% \lei{The meaning of this list would not be very clear to readers}
Every surviving render is then routed past a domain expert for visual sign-off, and only records passing all stages enter the release (17{,}900 as of May 2026). Figure~\ref{fig:dataset_overview} illustrates the pipeline (top) and the four downstream evaluation tasks (bottom).

\paragraph{Three released datasets.} \emph{BenchCAD} contains 17{,}900 verified CadQuery code parts and serves as the primary evaluation set. \emph{BenchCAD-QA} provides 2{,}400 paired image/code numeric-QA items. \emph{BenchCAD-Edit} provides 748 curated edit pairs across dimensional, additive, subtractive, and multi-step categories.
% \lei{Would make these into three sentences.}
Every release ships a Croissant 1.0 metadata file~\citep{Croissant2024} validated by the public checker, hosted on Hugging Face under \texttt{BenchCAD/BenchCAD} with verification-pipeline source under MIT and data under CC-BY-4.0. Full datasheet (Appendix~\ref{app:datasheet}) and per-family schemas accompany the release.

\subsection{Tasks}
\label{sec:tasks}

BenchCAD evaluates models on four tasks, each targeting a distinct capability: \textsc{Vision2Code} (end-to-end synthesis of executable code from images), 
\textsc{Edit Code} (instruction-following program editing), \textsc{Code QA} (symbolic understanding of CadQuery programs) and \textsc{Vision QA} (geometric reasoning from rendered views).

\paragraph{\textsc{Vision2Code}.} 
% \lei{Complete the sentence} 
Models generate executable CadQuery code from four canonical orthographic views. We use IoU between the rendered prediction and the ground-truth occupancy grid as the primary metric. We also report Chamfer distance for geometric error, feature score for fine CAD details, 

\paragraph{\textsc{Code Edit}.} Given an original CadQuery program and a natural-language edit instruction, the model must produce a minimally modified program whose rendered solid matches the target. Items cover five edit types T1--T5: literal replacement, chained transformation, relative computation, feature editing, and geometry rebuilding (Table~\ref{tab:task_taxonomy}). We use \texttt{Accuracy} as the headline metric, which measures headroom-normalised improvement over the original-to-target gap (Eq.~\eqref{eq:edit_acc}) and controls for varying per-pair difficulty.
% \lei{Same here, hard to understand without much context}
The remaining metrics help disambiguate ties and attribute failures to parametric exactness, geometric closeness, or syntactic execution.

\begin{figure*}[t]
  \centering
  \includegraphics[width=0.9\linewidth]{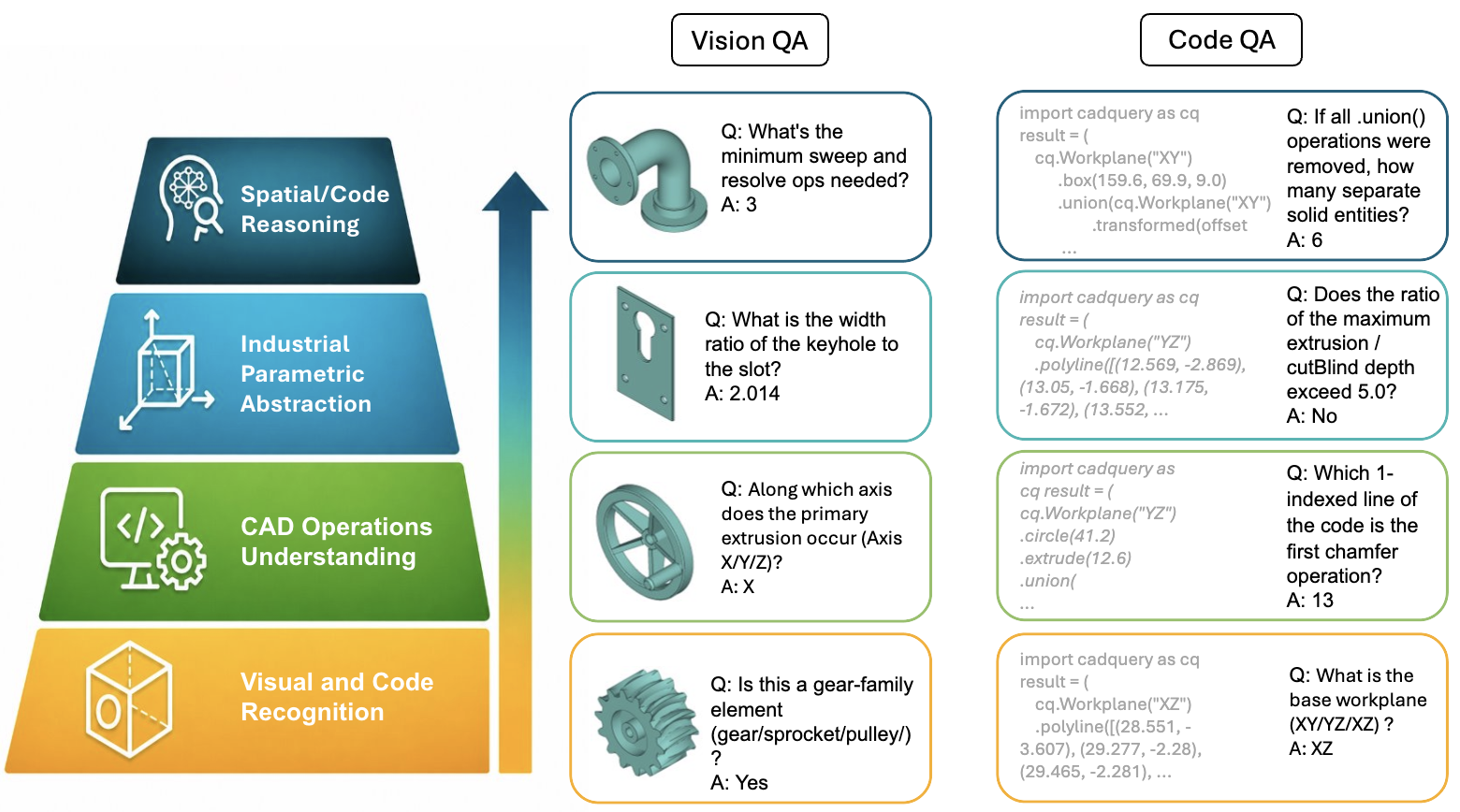}
  \caption{\textbf{BenchCAD-QA capability hierarchy.} Four-level capabilities (L1 Holistic Visual Recognition $\to$ L4 Spatial/Code Reasoning) with paired \textsc{Vision QA} / \textsc{Code QA} examples per level.}
  \label{fig:capability_hierarchy}

\end{figure*}
\paragraph{\textsc{Vision QA} and \textsc{Code QA}.}
\label{sec:capability}
We organise the QA bank along a four-level capability hierarchy (Figure~\ref{fig:capability_hierarchy}), from low-level perception at the base to high-level synthesis at the apex:
\emph{(i) Holistic Visual Recognition, $L_1$ } --- recognize the part family from multi-view renders and integrate views into a coherent volumetric understanding;
\emph{(ii) CAD Operations Understanding, $L_2$} --- read CadQuery (or infer from geometry) and map operations to features in the correct execution order;
\emph{(iii) Industrial Parametric Abstraction, $L_3$} --- abstract observed geometry into the parametric structure an engineer would write, respecting standard-table relations and engineering conventions;
\emph{(iv) Compositional Spatial-Code Reasoning, $L_4$} --- synthesize the preceding capabilities by planning spatially consistent CAD operations, maintaining dependencies across parameters and coordinate frames, and producing syntactically valid, executable parametric code.
% \lei{Find the following sentence hard to decipher}
The matched-pair design isolates the source of failure: a large gap between Vision QA and Code QA on identical questions indicates that errors arise primarily from visual recognition rather than reasoning over the queried attribute (the Holistic Spatial and Detailing Deficit, \S5.2), whereas low Code QA performance indicates a CAD-operation understanding bottleneck at $L_2$.

% \lei{hard to understand}
essential-op recall for key operation use, and execution rate for code validity. 

Question-bank construction, edit-pair protocol, rotation-invariant IoU, and scale-invariance ablation appear in Appendices~\ref{app:qbank}, \ref{app:edit_protocol}, \ref{app:iou}, and~\ref{app:ablation}.
\section{Experiments}
\label{sec:experiments}

\subsection{Setup}
\label{sec:exp_setup}
For each task, we include a \emph{blind baseline} that preserves the output format but removes the informative input: black views for \textsc{Image-to-Code}/\textsc{Image QA}, no code for \textsc{Code QA}, and unchanged original programs for \textsc{Code Edit}. These baselines measure gains beyond dataset priors and metric floors.
% \hc{provide more details}
and quantify each model's gain over it across all relevant metrics.

\subsection{Models}
\label{sec:exp_models}

Three model classes are evaluated:
\emph{(i) Frontier proprietary MLLMs:} GPT-4o~\citep{OpenAI2024GPT4o}, GPT-5.3 thinking / non-thinking~\citep{OpenAI2026GPT53Instant}, Claude Opus 4.7 thinking / non-thinking~\citep{Anthropic2026Claude47}, Gemini 3.1 Pro thinking / non-thinking~\citep{Google2026Gemini31}, OpenAI o3~\citep{OpenAI2025O3O4Mini}, Moonshot Kimi~\citep{MoonshotAI2025Kimi}.
\emph{(ii) Open-source MLLMs / code LLMs:} Qwen3-VL~\citep{Bai2025Qwen3VL}, InternVL3~\citep{Chen2025InternVL3}, gpt-oss-120b~\citep{OpenAI2025GPTOSS}, nemotron-3-super-120b-a12b~\citep{NVIDIA2026Nemotron}.
\emph{(iii) CAD-specialist lineage:} cadrille-RL~\citep{Kolodiazhnyi2026} and CADEvolve v3~\citep{Elistratov2026}. Closed models are queried through their official APIs; open-weights models via OpenRouter~\citep{OpenRouter2024} or the official repository.

\subsection{Evaluation}
\label{sec:exp_eval}
 For \textsc{Vision2Code} we report \texttt{exec\_pct}, mean voxel \texttt{IoU}, mean Chamfer distance \texttt{CD}, mean Hausdorff distance \texttt{HD}, Feature-F1, essential-op recall \texttt{ess}, and a composite \texttt{total} score (Table~\ref{tab:codegen_unified}). For \textsc{Edit Code}, with rendered solid $S_g$, original solid $S_o$, and target solid $S_t$, let $\mathrm{IoU}(S_a, S_b)$ denote the voxel intersection-over-union. Over $n$ edit records, we report normalized accuracy, which controls for pair-specific bias induced by varying original--target IoU:
\begin{align}
\label{eq:edit_acc}
\mathrm{Acc}_{\mathrm{norm}}
= \frac{1}{n}\sum_{i=1}^{n}
\mathrm{clip}\!\left(
\frac{\mathrm{IoU}(S_g^i, S_t^i) - \mathrm{IoU}(S_o^i, S_t^i)}
{1 - \mathrm{IoU}(S_o^i, S_t^i)}, 0, 1
\right).
\end{align}
Accuracy is a headroom-normalised improvement, where $1$ means the model fully traversed the original-to-target gap, and all records satisfies $\mathrm{IoU}(S_o^i, S_t^i)<0.99$. 
For Vision QA and Code QA (BenchCAD-QA, 2{,}400 paired image/code numeric items), accuracy is computed under $\pm 5\%$ tolerance for ratios and exact match for integers, broken out along the four-level capability hierarchy (\S\ref{sec:capability}).

\subsection{Training}
\label{sec:exp_training}
Beyond evaluation, we use BenchCAD as a training resource to test whether its operation breadth and standard-anchored families yield measurable capability gains, and whether these gains transfer beyond the trained families. We train an open Qwen3-VL-2B baseline reported in the bottom block of Table~\ref{tab:codegen_unified}, comparing matched-compute SFT and RL settings that differ only in training data composition. 
% \hc{Need to polish this paragraph, clearly describing the training methods}
\textbf{SFT.} We train three variants with the same optimizer, schedule, and held-out evaluation set: \emph{iid} uses BenchCAD plus extrusion-heavy auxiliary data, \emph{ood} follows the same recipe but removes 10 mechanical families, and \emph{baseline} uses no BenchCAD data.
\textbf{RL.} Starting from each SFT checkpoint, we apply an on-policy GRPO-style objective with reward $r{=}0.2\,\mathrm{ess}{+}0.8\,\mathrm{IoU}$ and $r{=}-1$ for parse errors outputs; the OOD-RL run also excludes the held-out OOD families. We report per-operation recall ablations in Table~\ref{tab:op_recall_train}, with full training details and curves in Appendix~\ref{app:training}.

\section{Results}
\label{sec:results}
\label{sec:exp_main}

\begin{figure*}[htbp]
  \centering
  \includegraphics[width=0.75\linewidth]{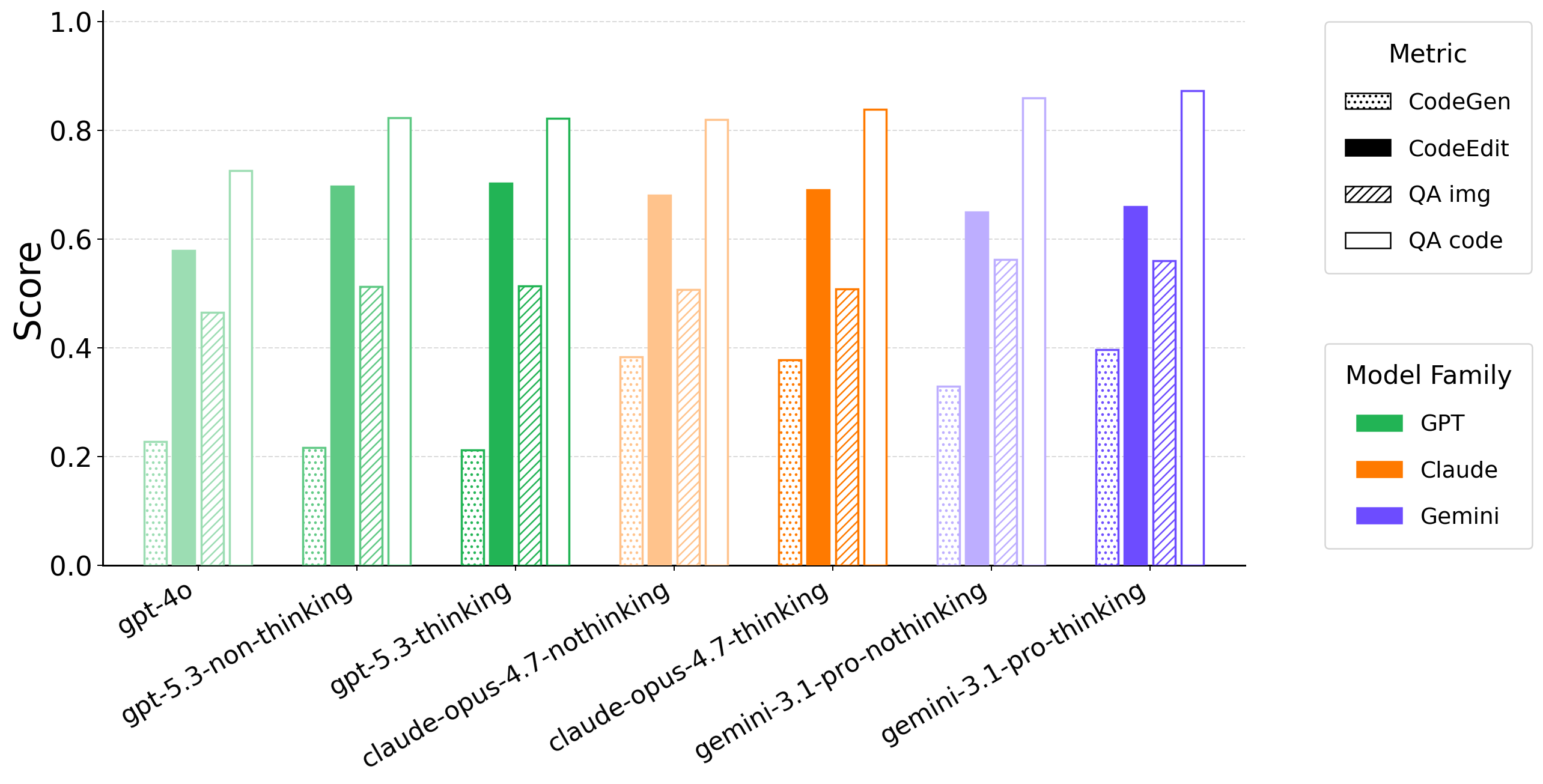}
  \caption{\textbf{Per-model performance across the four BenchCAD task categories}
  (frontier proprietary subset; open-source baselines are reported in Table~\ref{tab:main_eval_qa_img}). 
  Bar style encodes task, color encodes model family, and within-family variants distinguish thinking from non-thinking models. 
  Two patterns motivate our subsequent diagnostics: (i) QA-img consistently underperforms QA-code, highlighting the difficulty of holistic spatial recognition from visual evidence; and (ii) CodeEdit trails CodeGen across all models, indicating that instruction-guided modification of existing CAD programs remains harder than greenfield generation. 
  % \lei{The figure seems not in an aesthetic aspect ratio.}\hc{Agree; maybe use .pdf for higher resolution.}\hz{updated}
  } 
  \label{fig:overall_perform}
\end{figure*}

Our experiments provide a comprehensive and decoupled analysis of BenchCAD across four tasks (Figure~\ref{fig:overall_perform}). 
By separating results by modality, task type, and capability axis, and by tracing failures to shared reasoning bottlenecks, we show that BenchCAD is both a challenging evaluation benchmark and a structured training resource for improving CAD reasoning.

% \lei{Fig.~\ref{fig:overall_perform} is not discussed here. Also move the figure here}

\subsection{Overall Performance of 4 Tasks}
\label{sec:results_overall}

\begin{table*}[t]
\centering
\small
\setlength{\tabcolsep}{5pt}
\renewcommand{\arraystretch}{1.15}
\caption{\textbf{Main evaluation on BenchCAD-QA (\texttt{qa\_img} modality), by capability axis.} Given multi-view renders, models answer visual CAD questions; closed-source MLLMs outperform the open-source MLLMs but remain weak on operation understanding and parametric abstraction.}
% \lei{Briefly explain the task and summarize the results here in the caption. Would be good to have up arrows for the header.}}
\label{tab:main_eval_qa_img}
\begin{tabular}{l c c c c c}
\toprule
Model
 & \shortstack{Holistic Visual\\Recognition}
 & \shortstack{CAD Operation\\Understanding}
 & \shortstack{Industrial Parametric\\Abstraction}
 & \shortstack{Spatial\\Reasoning}
 & \shortstack{Total\\score} \\
\midrule
opus-4.7         & 0.699 & 0.464 & 0.426 & 0.668 & 0.526 \\
opus-4.7-thinking            & 0.715 & \textbf{0.485} & 0.421 & 0.614 & 0.530 \\
gemini-3.1-pro  & \textbf{0.750} & 0.462 & 0.536 & \textbf{0.688} & \textbf{0.587} \\
gemini-3.1-pro-thinking     & 0.722 & 0.426 & \textbf{0.551} & 0.669 & 0.576 \\
gpt-4o                              & 0.599 & 0.408 & 0.431 & 0.396 & 0.464 \\
gpt-5.3-chat-latest                 & 0.636 & 0.423 & 0.488 & 0.548 & 0.513 \\
gpt-5.3-thinking                    & 0.650 & 0.429 & 0.482 & 0.534 & 0.514 \\
moonshot-v1-128k     & 0.556 & 0.387 & 0.427 & 0.334 & 0.442 \\

  moonshot-v1-8k-vision         & 0.600 & 0.246 & 0.465 & 0.181 & 0.447 \\                                                                    
  openai-o3                             & 0.328 & 0.188 & 0.398 & 0.560 & 0.327 \\  
% google-gemma-4-31b-it       & --    & --    & --    & --    & --    \\
% kimi-k2.6                           & --    & --    & --    & --    & --    \\
\midrule
\textit{blank image (baseline)}     & 0.376    & 0.325    & 0.418    & 0.296    & 0.375    \\
\bottomrule
\end{tabular}
\end{table*}

\paragraph{\textsc{Vision2Code}.} The unified leaderboard (Table~\ref{tab:codegen_unified}, Appendix~\ref{app:full_results}) shows two patterns:
(i) the specialist CAD lineage transferred from DeepCAD scores well on IoU but underperforms on non-extrude operations;
(ii) frontier MLLMs cap mid-range (gemini-3.1-pro-thinking total 0.318) and exhibit non-monotonic thinking-mode behaviour (Table~\ref{tab:ablation_img_size}).

\begin{figure}[t]
  \centering
  \begin{minipage}{0.6\linewidth}
    \centering
    \includegraphics[width=\linewidth]{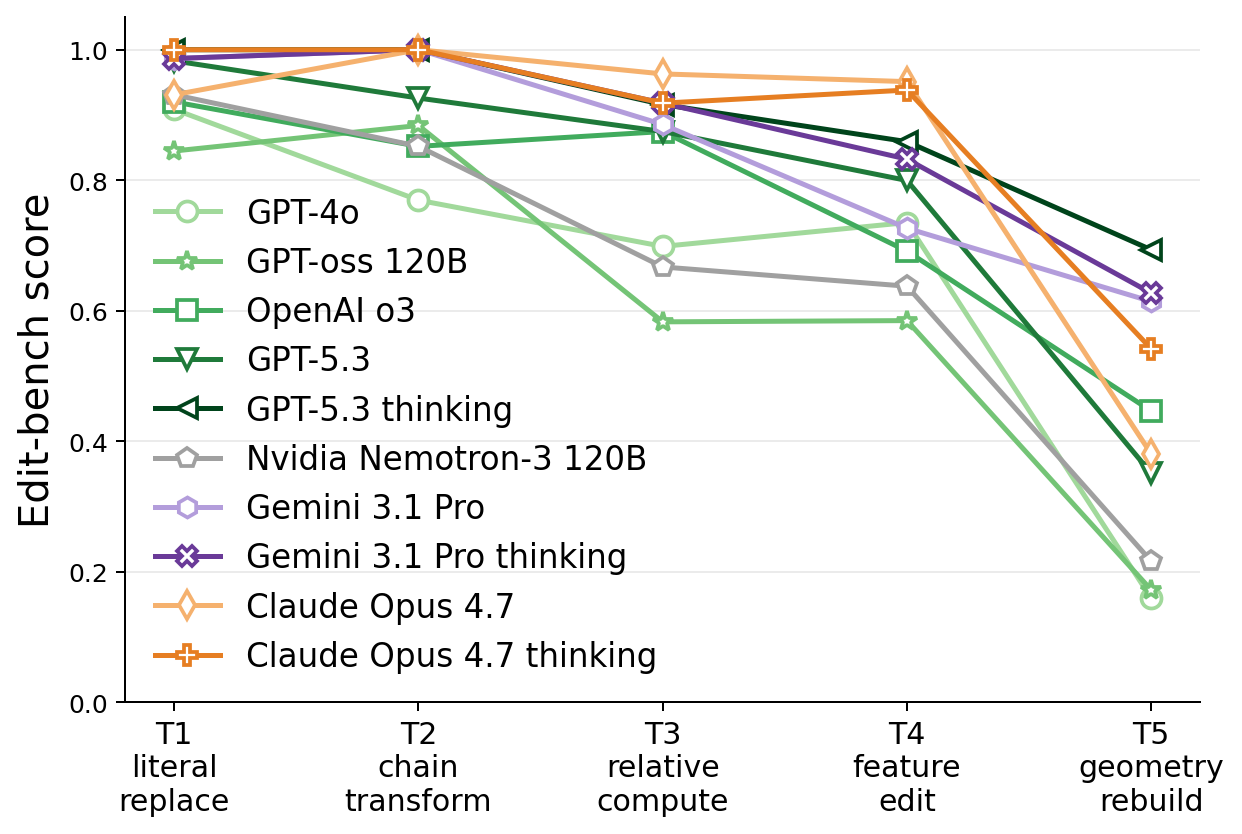}
    \captionof{figure}{\textbf{BenchCAD-Edit by task type.}}
    \label{fig:edit_task}
  \end{minipage}\hfill
  \begin{minipage}{0.4\linewidth}
    \centering
    \captionof{table}{\textbf{BenchCAD-Edit Accuracy.}}
    \label{tab:edit_leaderboard}
    \small
\setlength{\tabcolsep}{3pt}
\begin{tabular}{@{}lccc@{}}
\toprule
\textbf{Model}& \textbf{Thinking} & \textbf{Acc.} \\
\midrule
GPT-5.3            & \cmark & \textbf{0.865} \\
Claude Opus 4.7    & \cmark & 0.853 \\
Gemini 3.1 Pro     & \cmark & 0.837 \\
Claude Opus 4.7    & --     & 0.811 \\
Gemini 3.1 Pro     & --     & 0.795 \\
GPT-5.3            & --     & 0.740 \\
OpenAI o3          & \cmark & 0.708 \\
GPT-4o             & --     & 0.615 \\
Nemotron-3 120B    & --     & 0.608 \\
GPT-oss 120B       & --     & 0.561 \\
Baseline (no change) & --     & 0.000 \\
\bottomrule
\end{tabular}
  \end{minipage}
\end{figure}

\paragraph{\textsc{Code Edit}.} 
%\paragraph{Code editing.}
Table~\ref{tab:edit_leaderboard} reports the BenchCAD-Edit leaderboard using mean-normalized improvement as the headline metric, which discounts the varying difficulty of each edit pair. Overall scores show a clear gap between model families, but the per-type breakdown in Fig.~\ref{fig:edit_task} reveals a more diagnostic pattern: simple API-level edits are nearly solved by modern models, while compositional edits remain difficult. Replacing the textual instruction with a target render collapses every model to near-zero, since a render specifies geometry but not numbers; augmenting the text instruction with the original four-view image barely moves the aggregate and even hurts the strongest thinking model (Fig.\ref{fig:edit_image_type}), indicating that the textual instruction does the heavy lifting and the visual signal is at best a clarifier. The error analysis below ties each regime to specific failure modes; per-model F-code distributions, the full L1--L4 attribution, and the image-conditioned variant are deferred to Appendix~\ref{app:edit_protocol}.

\paragraph{\textsc{Vision QA} and \textsc{Code QA}.}
Table~\ref{tab:main_eval_qa_img} shows that Vision QA remains challenging for current frontier MLLMs: models can often recognize global part shapes, but struggle to infer the underlying CAD operations and parametric design intent. Direct access to CadQuery code substantially improves performance, with the best Code QA models reaching total scores around 0.838 compared with 0.587 for Vision QA. This modality gap indicates that explicit programs make geometric and parametric information easier to extract than rendered images. However, Spatial / Code Reasoning remains weaker even with code access, showing that the benchmark still requires precise spatial and operational reasoning beyond surface-level parsing. Full Code QA results are reported in Appendix~\ref{app:full_results}.

\subsection{Error Analysis}
\label{sec:results_failure_analysis}
\label{sec:exp_failure_analysis}

Across tasks, we find that many failures share common causes. Most can be attributed to deficiencies in three core capabilities.

\textbf{Holistic Spatial Recognition.}
Visual recognition failures occur along two axes: detailed feature counting and spatial grounding.
%
% For instance, 4 out of 8 models fail to identify the keyway adjacent to the central hole in a gear, as the two features are visually blended together in the rendered view.

\begin{wrapfigure}{r}{0.35\linewidth}
  \vspace{-\baselineskip}
  \centering
  \includegraphics[width=\linewidth]{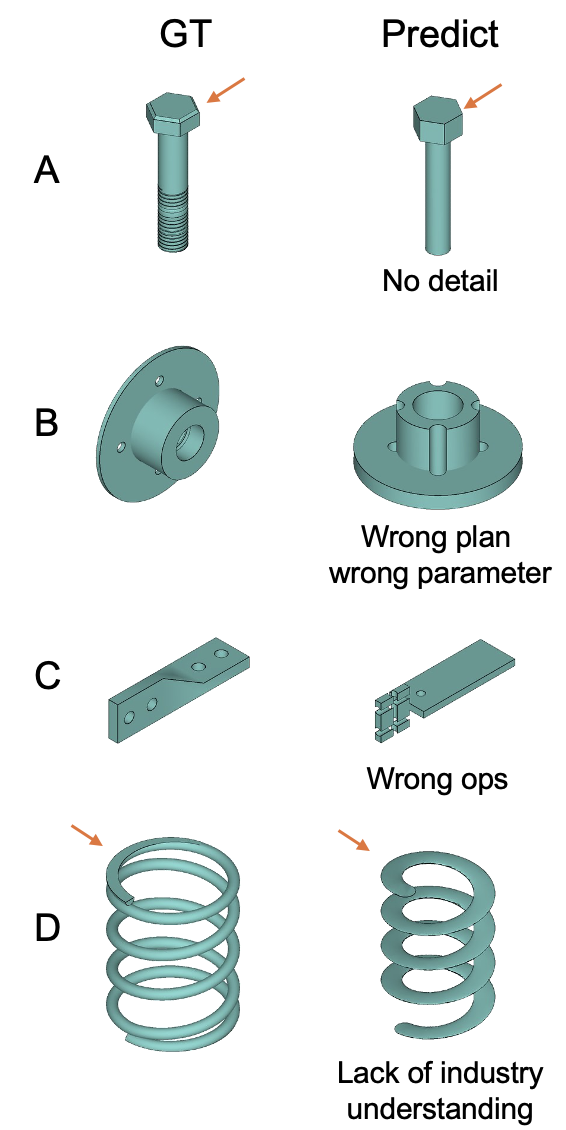}
  \caption{\textbf{Examples of failures in codegen.} Zoom-in for more details.}
  \label{fig:err_analysis_codegen}
  \vspace{-0.5\baselineskip}
\end{wrapfigure}

A hex-head bolt (Figure~\ref{fig:err_analysis_codegen}A) illustrates a fine-detail failure. 
GPT-5.3 recovers the overall bolt body, but fails to capture fine-grained details such as the thread structure and the chamfer on the top face. 
The generated part is therefore visually similar at a coarse level but misses small, human-obvious CAD features. Consistently, GPT-5.3-chat fails to identify the number of chamfer operations in the corresponding vision QA ($L_1$) on chamfer number which indicates an visual recognition problem.

A bearing retainer cap (Figure~\ref{fig:err_analysis_codegen}B) exposes a global spatial-frame failure. 
The model captures several fine-grained features but fails to infer the correct extrusion direction: the generated code extrudes from the XY plane rather than the target XZ plane. This error is common for geometries generated based on XZ and YZ workplanes. It is partially diagnosed by the substantially higher 24-axis rotation-invariant IoU compared with the single-axis IoU, suggesting that the predicted shape is geometrically similar but expressed in the wrong spatial frame (Appendix~\ref{app:iou}). 
The same example also reveals parametric-abstraction failures across multiple features.

\textbf{Operation understanding.} A twisted bracket (Figure~\ref{fig:err_analysis_codegen}C) is generated as two mutually perpendicular brackets without twisted connection; the requisite twist-extrusion is absent from the emitted program entirely. 
The model recognizes the holistic spatial (L1) but fails to map the visible torsion to the corresponding CAD operation, exposing an Op Vocabulary Gap (Appendix~\ref{app:op_gap_full}) at the operation-understanding layer.

\textbf{Industrial Parametric abstraction.}
A standardized DIN~2095 coil spring (Figure~\ref{fig:err_analysis_codegen}D) exposes a parametric-abstraction failure. 
The standard requires closed and ground ends: the coil pitch is locally reduced near both terminals, causing the end turns to flatten into planar bearing surfaces. 
The model recognizes the spring and the coil pitch is not a circle, and emits a helical sweep, but collapses this end-specific pitch variation into a uniform helix with an incorrect cross-section.
It therefore captures the coarse family and operation (L1 \& L2) while missing the standard-driven parameterization that an engineer would explicitly encode. 
This failure highlights why BenchCAD anchors its families to engineering standards: the benchmark tests not only whether models can name a part or invoke the right CAD operation, but whether they can recover the industrialized structured design rules underlying real parametric CAD.

Together, these cases show how our layered framework turns \textsc{Vision2Code} errors from an opaque ``wrong code'' label into level-specific diagnostics: the bracket fails at $L_2$, while the spring succeeds at $L_1$--$L_2$ but fails at $L_3$. 
Without per-level attribution these distinctions collapse into a single low IoU score, and the corresponding improving signal --- which level to strengthen --- is lost.

\subsection{BenchCAD as a Training Resource for Improving Model Capabilities}
\label{sec:results_training}

BenchCAD is not only an evaluation benchmark but also a useful training resource. Our Qwen3-VL-2B trained on BenchCAD achieves the best in-distribution code-generation result, with a CodeGen score of 0.7631 in Table~\ref{tab:codegen_unified}. Even when 10 mechanical families are held out, IID-SFT+RL transfers non-trivially to unseen OOD families, suggesting reusable CAD priors beyond family-level memorization. BenchCAD training also broadens the generated operation vocabulary, improving use of advanced operations such as \texttt{revolve}, \texttt{sweep}, \texttt{loft}, and \texttt{fillet} (Table~\ref{tab:op_recall_train}). However, the substantial IID/OOD gap shows that generalization to novel mechanical families remains an open challenge (Appendix~\ref{app:training}).
% \hc{This paragraph reads a bit rushed; maybe add more detail.}

\section{Conclusion}
\label{sec:conclusion}

BenchCAD provides a capability-decomposed benchmark for industrial CAD reasoning, moving evaluation beyond rendered shape similarity toward the operations, parameters, constraints, and editable program structure that define practical CAD models. 
Across 10+ frontier and CAD-specialized models, BenchCAD shows that current systems often recover coarse geometry but remain unreliable at fine spatial grounding, CAD-operation selection, industrial parametric abstraction, and localized program editing. 
These failures appear consistently across paired image/code QA, operation-rich Vision2Code generation, and verified edit tasks, revealing why visually plausible outputs can still fail as engineering CAD programs.
As a training source, BenchCAD improves rare-operation recall and yields partial transfer to held-out families; however, the persistent OOD gap shows that robust industrial CAD generalization remains difficult.
We release BenchCAD and its Croissant 1.0 metadata under CC-BY-4.0 on Hugging Face, together with the evaluation harness under the MIT license on GitHub, to support more diagnostic progress toward industry-grade CAD automation.

% \begin{ack}
% \textcolor{red}{[Anonymised for submission. Restore acknowledgments and funding disclosure for camera-ready.]}
% \end{ack}

\bibliographystyle{plainnat}
\bibliography{refs}

% --- appendix ---
\appendix
\newpage
\section{Limitations}
\label{app:limitations}

\paragraph{Standard-parametric coverage is not full industrial CAD.} BenchCAD parts are generated from expert-authored standard-parametric families, not extracted from real engineering repositories.
We mitigate this by (i) anchoring 49\% of families to published ISO/DIN/EN/ASME/IEC specification tables and (ii) including verified Fusion360 Gallery and DeepCAD subsets for OOD comparison, but our parts do not capture the full diversity of proprietary industrial design (e.g.\ hand-drawn manufacturing tolerances, undocumented design intent, assembly information).

\paragraph{Standard-anchored does not mean standard-compliant.} A family declared as \texttt{standard = "ISO 23509"} samples within ISO parameter ranges and enforces inter-parameter relations from the standard, but does not validate against the full set of manufacturing tolerances or material specifications mandated by that standard.

\paragraph{Industrial Common Sense Gap measurement.} Our non-target preservation metric uses voxel IoU on the spatial complement of the targeted feature; this captures most leakage modes but can miss small parametric shifts whose voxel footprint is subthreshold.
\S\ref{app:edit_protocol} discusses the protocol, including a complementary -AST diff metric we report alongside.

\section{Extended Related Work: CAD Code-Generation and Editing Benchmarks}
\label{app:related_extended}

\paragraph{Code generation.}
DeepCAD~\citep{Wu2021} introduces a 178K-model corpus of Onshape parts encoded as discrete operation tokens; Text2CAD~\citep{Khan2024} extends DeepCAD with 660K natural-language annotations across four abstraction levels. CAD-Coder~\citep{Guan2025} reformulates the same data as CadQuery Python source and applies SFT$+$GRPO with chain-of-thought and Chamfer reward (mean CD $6.54{\times}10^{-3}$ on Text2CAD). The CAD-Recode~\citep{Rukhovich2025}/cadrille~\citep{Kolodiazhnyi2026}/CADEvolve~\citep{Elistratov2026} lineage shares a Qwen2-VL-2B backbone: CAD-Recode fine-tunes on 1M sketch-and-extrude scripts; cadrille adds multi-modal inputs and Dr.CPPO online RL with piecewise IoU reward (DeepCAD IoU 92.2, Fusion360 IoU 84.6, 0.0\% invalidity); CADEvolve expands the corpus to $\sim$1.3M scripts via an LLM-driven evolutionary loop covering extrude/revolve/loft/sweep/fillet/chamfer/shell/Boolean/patterns --- broadening the operation distribution but still without helical sweeps or parametric involute-gear construction. The CADEvolve Hugging Face release contains only sentence embeddings (\texttt{.npy}), not the source code, so its full operation surface is not directly auditable; the public GitHub repo ships only $46$ hand-written seed programs.

\paragraph{Editing.}
DeepCAD~\citep{Wu2021} introduces a 178K-model corpus of Onshape parts encoded as discrete operation tokens; Text2CAD~\citep{Khan2024} extends DeepCAD with 660K natural-language annotations across four abstraction levels. CAD-Coder~\citep{Guan2025} reformulates the same data as CadQuery Python source and applies SFT$+$GRPO with chain-of-thought and Chamfer reward (mean CD $6.54{\times}10^{-3}$ on Text2CAD). The CAD-Recode~\citep{Rukhovich2025}/cadrille~\citep{Kolodiazhnyi2026}/CADEvolve~\citep{Elistratov2026} lineage shares a Qwen2-VL-2B~\citep{Qwen2VL} backbone: CAD-Recode fine-tunes on 1M sketch-and-extrude scripts; cadrille adds multi-modal inputs and Dr.CPPO online RL with piecewise IoU reward (DeepCAD IoU 92.2, Fusion360 IoU 84.6, 0.0\% invalidity); CADEvolve expands the corpus to $\sim$1.3M scripts via an LLM-driven evolutionary loop covering extrude/revolve/loft/sweep/fillet/chamfer/shell/Boolean/patterns --- broadening the operation distribution but still without helical sweeps or parametric involute-gear construction. The CADEvolve Hugging Face release contains only sentence embeddings (\texttt{.npy}), not the source code, so its full operation surface is not directly auditable; the public GitHub repo ships only $46$ hand-written seed programs.

\section{Full Comparison Against Prior CAD Code-Generation Work}
\label{app:cmp_prior}

Table~\ref{tab:cmp_prior} provides the full per-axis comparison referenced in \S\ref{sec:related}, including all evaluation-side properties (\#Industrial families, \#Std codes, \#Distinct CQ ops, Adv ops, Edit / QA tasks, \#Eval models) across DeepCAD, Fusion360 Gallery, Text2CAD, CADPrompt, CAD-Coder, CAD-Recode, cadrille, and CADEvolve.

\begin{table*}[t]
\centering
\small
\setlength{\tabcolsep}{5pt}
\caption{\textbf{BenchCAD versus prior CAD code-generation work.} BenchCAD is a \emph{unified, capability-decomposed evaluation framework} for large-model CAD reasoning, distinct in purpose from the CadQuery-VLM lineage (CAD-Recode, cadrille, CADEvolve), which releases training corpora alongside small fine-tuned models (Qwen 1.5B--2B) saturating $\sim$92\% IoU on legacy sketch+extrude splits. \textbf{Type:} \emph{C+M} = corpus+model, \emph{D} = dataset, \emph{B} = benchmark. \textbf{\#Fam.} = named industrial part families. \textbf{\#Std} = distinct ISO/DIN/EN/ASME/IEC specification codes. \textbf{Op surface}: \emph{narrow} = sketch+extrude IR or a small CadQuery subset; \emph{broad} = wide CadQuery API including advanced solid ops (precise per-corpus counts and protocol in App.~\ref{app:ops}). \textbf{Adv} = all four advanced operation families (helix, loft/sweep, twist-extrude, parametric involute-gear) exercised. $\checkmark$ = supported, p.\ = partial, -- = absent.}
\label{tab:cmp_prior}
\begin{tabular}{lccrrrlccc}
\toprule
Benchmark & Yr & Type & \#Sam. & \#Fam. & \#Std & Op surface & Adv & Edit & QA \\
\midrule
DeepCAD~\citep{Wu2021}                   & '21 & C+M & 178K   & -- & 0  & narrow & --  & --   & --   \\
Fusion360 G.~\citep{Willis2021fusion}    & '21 & D   & 8K     & -- & 0  & narrow & --  & --   & --   \\
Text2CAD~\citep{Khan2024}                & '24 & C+M & 170K   & -- & 0  & narrow & --  & --   & --   \\
CADPrompt~\citep{Alrashedy2025}          & '25 & B   & 200    & -- & 0  & broad  & p.\ & --   & --   \\
CAD-Coder~\citep{Guan2025}               & '25 & C+M & 163K   & -- & 0  & narrow & --  & p.\  & --   \\
CAD-Recode~\citep{Rukhovich2025}         & '25 & C+M & 1M+7K  & -- & 0  & narrow & --  & p.\  & seq.\\
cadrille~\citep{Kolodiazhnyi2026}        & '26 & C+M & 1M+7K  & -- & 0  & narrow & --  & --   & --   \\
CADEvolve~\citep{Elistratov2026}         & '26 & C+M & 1.3M   & -- & 0  & broad  & p.\ & --   & --   \\
\midrule
\textbf{BenchCAD (ours)} & \textbf{'26} & \textbf{D+B} & \textbf{17{,}900} & \textbf{106} & \textbf{43} & \textbf{broad}+ & $\checkmark$ & $\checkmark$ & $\checkmark$ \\
\bottomrule
\end{tabular}
\end{table*}

\section{Standard Codes Covered}
\label{app:standards}
Table~\ref{tab:standards} reports the standards grounding of BenchCAD. 
Nearly half of all families, 52/106 (49\%), are tied to ISO, DIN, EN, ASME, or IEC specification tables, so their dimensions are sampled from real engineering ranges rather than arbitrary shapes. 
The remaining 54 custom families cover bespoke industrial parts without a formal standard, while still using analogous proportional design rules.

\begin{table}[t]
\centering
\small
\caption{\textbf{Standard-anchored families.} 49\% of BenchCAD families (52/106) bind their parameters to ISO/DIN/EN/ASME/IEC specification tables, sampling at values drawn from real engineering ranges rather than arbitrary geometry. The bottom row counts the 54 custom families covering bespoke industrial parts (twisted brackets, lobed knobs, etc.) governed by analogous proportional rules without formal standards.}
\label{tab:standards}
\begin{tabular}{llrr}
\toprule
Standard & Codes covered (sample) & \#Codes & \#Fam. \\
\midrule
ISO   & 22, 53, 113, 272, 606, 1234, 2339 \ldots & 17 & 18 \\
DIN   & 315, 338, 471/472, 580, 660, 705, 950, 2095, \ldots & 21 & 21 \\
EN    & 10034, 10056, 10219, 10279                       & 4 & 5 \\
ASME  & B1.20.1, B16.5, B16.9                            & 3 & 5 \\
IEC   & 60072-1, 60086                                   & 2 & 2 \\
\midrule
\textbf{Total standard-anchored}   & & \textbf{47} & \textbf{52 (49\%)} \\
Custom (no formal standard)        & & --          & 54 \\
\bottomrule
\end{tabular}
\end{table}

\section{Dataset Generation Details}
\label{app:datagen}

\paragraph{Subfamilies.} A family may further branch into one or more \emph{subfamilies} that capture mating, assembly, or construction variants of the same part type --- for instance, a threaded-fastener family may split into male (\texttt{bolt}) and female (\texttt{nut}) subfamilies that share the same thread/pitch parameter table but differ in build chain; a retaining-ring family may split into internal (groove inside a bore) and external (groove on a shaft) variants; a thread family may split into single- and multi-start helices. Subfamilies reuse their parent family's parameter schema and standard-table anchoring, but specify their own deterministic builder. They are sampled and rendered independently, so a single named family in the per-family analysis can contribute several subfamily$\times$tier buckets to the verified release.

\paragraph{Difficulty tiers.} Each (sub)family exposes three tiers --- \emph{easy} / \emph{medium} / \emph{hard} --- defined by parameter complexity. Easy uses default parameter ranges with optional features disabled (e.g., a coil spring with constant pitch and no end treatment). Medium expands parameter ranges and enables a subset of optional features. Hard activates all optional features and uses extreme but still standard-compliant parameter ranges (e.g., variable pitch with closed-and-ground spring ends), exercising the more advanced construction operations of the family. Sampling is balanced approximately uniformly across tiers within each (sub)family.

\paragraph{Sandbox failure-mode taxonomy.} Each generated CadQuery program is executed in a sandbox subprocess; records hitting any of the following are quarantined and excluded from the release: (i) \emph{parse / import error} --- the program fails Python parsing or CadQuery API resolution; (ii) \emph{runtime exception} --- a CadQuery call raises (e.g., on a non-positive radius or an empty selector); (iii) \emph{timeout} --- execution exceeds a 30\,s wall-clock budget, typically caused by pathological boolean or sweep construction; (iv) \emph{degenerate volume} --- the resulting solid has $\leq 10^{-6}$\,mm$^3$ volume or an inverted (negative-determinant) transformation chain. Programs surviving all four checks are rendered into multi-view images and routed past a domain expert for visual sign-off; only records passing every stage enter the release.

\section{Model Training}
\label{app:training}
\paragraph{SFT training mixture.}
(1) \textbf{iid}: BenchCAD (all 106 families) + extrusion-heavy data (text2cad, cad-recode);
(2) \textbf{ood}: same recipe, 10 mechanical families held out;
(3) \textbf{baseline}: text2cad + cad-recode only (no BenchCAD).
Mixing 33\% BenchCAD / 67\% HQ for (1)/(2); 100\% HQ for (3). Backbone Qwen3-VL-2B.

\paragraph{OOD holdout selection.}
We hold out 10 CAD families sampled uniformly at random from BenchCAD families that are sufficiently represented and the operations are covered in the iid but not in the baseline.

\begin{figure*}[t]
  \centering
  \includegraphics[width=0.85\linewidth]{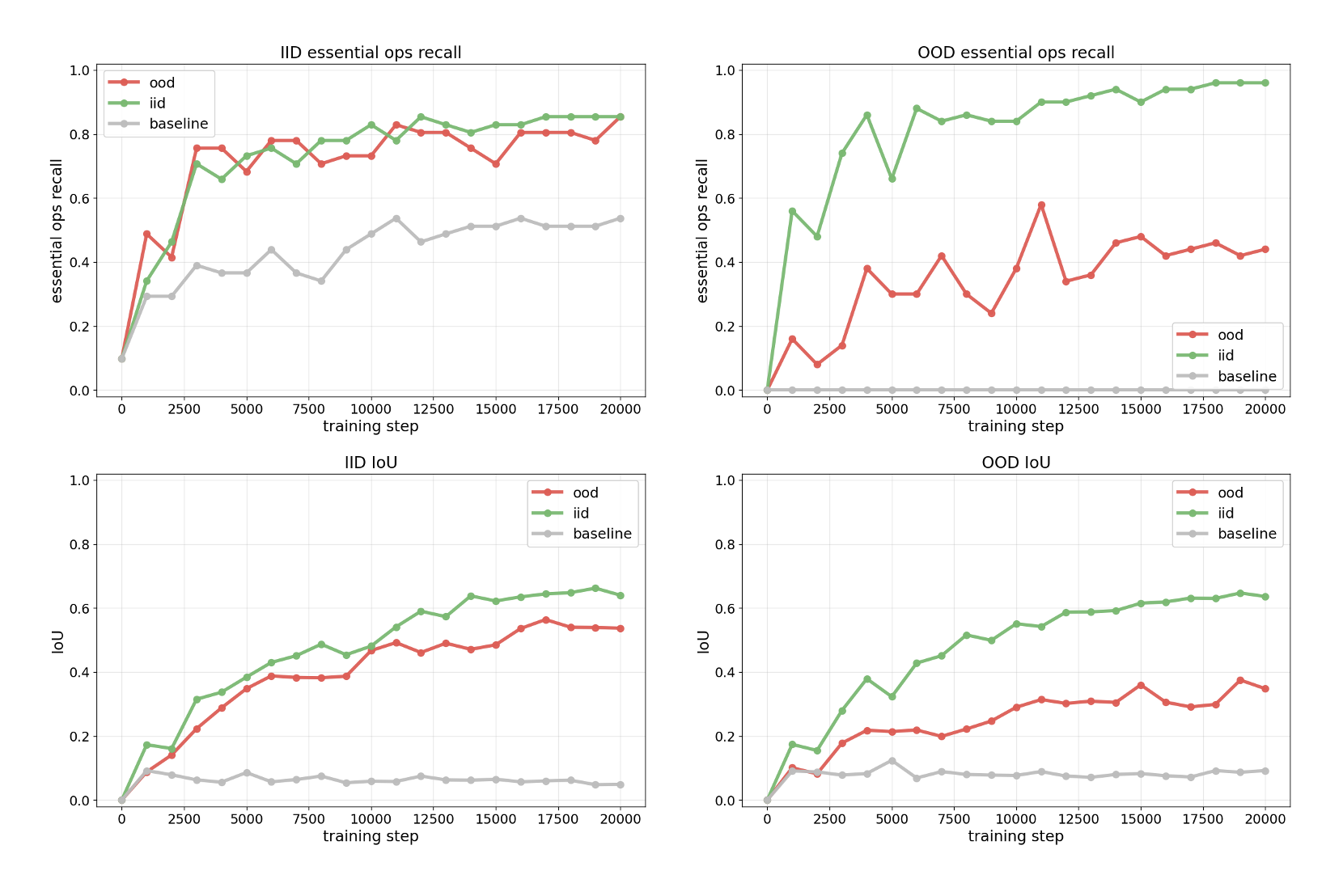}
  \caption{\textbf{The generalisation gap on BenchCAD.} Qwen3-VL-2B trained on three data mixtures, evaluated on the BenchCAD validation set throughout training.
    \emph{(a) OOD IoU vs.\ training step.} The IID-trained run (green) climbs highest; the OOD run (red, trained without the held-out family slice) plateaus mid-range; the baseline (grey, no BenchCAD) stays near the floor.
    \emph{(b) OOD essential-op pass rate.} IID reaches the highest rate, OOD plateaus mid-range, baseline stays at $\sim$0\% --- the \textbf{CAD Operational Blindspot}.
    The deficit is not optimisation budget, not backbone capacity, and not training-corpus operation coverage alone --- it is the fundamental difficulty of constructing industry-grade parametric CAD code on \emph{unseen} family distributions.}
  \label{fig:generalisation_gap}
\end{figure*}

\paragraph{SFT training setup.}
We fine-tune Qwen3-VL-2B-Instruct~\citep{Bai2025Qwen3VL} under a standard supervised code-generation setup. Inputs consist of one rendered view and a textual instruction, and targets are CadQuery programs.We use AdamW with learning rate $2{\times}10^{-4}$, cosine decay, $2$k warmup steps, weight decay $0.01$ with bf16 precision. The training evaluation curve (n=30) shows significant gap between model with OOD holdouts and model trained on all family (Fig~\ref{fig:generalisation_gap}).

\paragraph{RL training setup.}
We initialize RL from the matching SFT checkpoint: IID-SFT uses the full BenchCAD training set, whereas OOD-SFT excludes the held-out OOD families. RL is trained with a top-$N$ GRPO-style objective without advantage normalization on a mixed prompt pool from BenchCAD, DeepCAD, and Fusion360; in the OOD setting, the held-out families remain excluded. Rewards combine geometric fidelity with an essential-operation term when available. We use learning rate $2{\times}10^{-5}$, 16 rollouts per prompt, and batch size 128.

\begin{table}[t]
\centering
\small
\setlength{\tabcolsep}{5pt}
\renewcommand{\arraystretch}{1.08}
\caption{\textbf{IID--OOD generalization under SFT and RL.}
We report geometric fidelity, execution rate, essential-operation score, and the final score. The 106 families are separated into two family group IID and OOD. The iid-rl(D) trained on all dataset (IID + OOD) demonstrates the highest performances on both IID and OOD family.}
\label{tab:iid_ood}
\begin{tabular}{@{}llccccc@{}}
\toprule
Checkpoint & Split & IoU$\uparrow$ & Exec.$\uparrow$ & Ess.$\uparrow$ & Score$\uparrow$ \\
\midrule
\multirow{2}{*}{ood-sft (A)}
& IID  & 0.577 & 94.6\% & 0.789 & 0.6064 \\
& OOD  & 0.397 & 89.8\% & 0.305 & 0.3776 \\
\midrule
\multirow{2}{*}{ood-rl (B)}
& IID & 0.757 & 99.1\%  & 0.865 & 0.7467 \\
& OOD  & 0.464 & \textbf{100.0\%} & 0.339 & 0.4604 \\
\midrule
\multirow{2}{*}{iid-sft (C)}
& IID & 0.640 & 87.3\% & 0.857 &  0.6225 \\
& OOD  & \ 0.635 & 96.6\% & 0.983 & 0.6920 \\
\midrule
\multirow{2}{*}{iid-rl (D)}
& IID & 0.753 & 98.9\% & 0.883 & 0.7635 \\
& OOD & \textbf{0.761} & 99.6\% & \textbf{0.989} & \textbf{0.8214} $\star$ \\
\bottomrule
\end{tabular}
\end{table}
\paragraph{Model Generalization.}
We report the full training results in Table~\ref{tab:iid_ood} and the corresponding operation-level recall in Table~\ref{tab:op_recall_train}. 
Compared with the pretrained baseline, both IID- and OOD-family SFT substantially improve the use of advanced CadQuery operations, indicating that BenchCAD provides effective supervision for operation-level CAD synthesis. 
Applying RL on top of the OOD-SFT checkpoint further improves both IID and held-out OOD performance, suggesting that geometry-based optimization improves executable fidelity beyond supervised imitation. 
However, the IID-trained SFT model obtains the strongest OOD score overall, indicating that broad family coverage during supervised training remains a key driver of generalization to unseen mechanical designs.

\section{Full Per-Model Per-Task Results}
\label{app:full_results}

Table~\ref{tab:dataset} reports the dataset composition; Table~\ref{tab:capability} maps each task to its sub-capability subset; Tables~\ref{tab:main_eval_qa_img} and~\ref{tab:main_eval_qa_code} report the per-model QA leaderboard (Vision QA / Code QA); Table~\ref{tab:family_cliff} stratifies \texttt{img2cq} by family difficulty. The unified Vision2Code leaderboard (Table~\ref{tab:codegen_unified}) is in the main body.

\begin{table}[t]
\centering
\small
\caption{\textbf{BenchCAD dataset composition.} Three released datasets: the verified CadQuery code core (17{,}900 parts spanning 106 families, each with parametric source, executable STEP, four canonical-view renders, parameter JSON, and op list), the paired QA bank (2{,}400 questions evaluated under both visual and code conditioning), and the Edit subset (748 before/after pairs).}
\label{tab:dataset}
\begin{tabular}{lrrl}
\toprule
Dataset & \#Records & \#Families & GT artefacts \\
\midrule
BenchCAD                     & 17{,}900             & 106 & code $\cdot$ STEP $\cdot$ 4 views $\cdot$ params $\cdot$ ops \\
BenchCAD-QA (paired img/code)& 2{,}400 $\times 2$   & 106 & question $+$ gold answer \\
BenchCAD-Edit                & 748                  & 106 & before/after $+$ NL instruction \\
\bottomrule
\end{tabular}
\end{table}

\begin{table}[t]
\centering
\small
\caption{\textbf{Capability decomposition.} BenchCAD's five tasks exercise known subsets of the four-level capability hierarchy: $L_1$ Holistic Visual and Cadquery Code Recognition (recognise + integrate multi-view), $L_2$ CAD Operations Understanding (understanding code, map ops to features), $L_3$ Industrial Parametric Abstraction (parameter structure, standard conventions from industry domain knowledge), and $L_4$ Spatial/Code Reasoning (compose all into executable program). Score differences across paired tasks (e.g.\ \texttt{qa\_img}$-$\texttt{qa\_code}) directly diagnose which level is the bottleneck. ``part.'' indicates partial coverage.}
\label{tab:capability}
\begin{tabular}{llccccc}
\toprule
Task & Input $\to$ Output & $L_1$ Vis & $L_1$ Code & $L_2$ Op & $L_3$ Param & $L_4$ Synth \\
\midrule
\texttt{img2cq}     & 4 views $\to$ code                  & \cmark & - & \cmark & \cmark & part. \\
\texttt{qa\_img}    & 4 views $+$ Q $\to$ text          & \cmark & - & \cmark & \cmark & \cmark \\
\texttt{qa\_code}   & code $+$ Q $\to$ text             &  -     & \cmark & \cmark & \cmark & \cmark \\
\texttt{edit\_img}  & views $+$ code $\to$ code  & \cmark & \cmark & \cmark & \cmark & \cmark \\
\texttt{edit\_code} & code $+$ instr $\to$ code            &   -   & \cmark & \cmark & \cmark & part. \\
\bottomrule
\end{tabular}
\end{table}

\begin{table*}[t]
\centering
\small
\setlength{\tabcolsep}{5pt}
\renewcommand{\arraystretch}{1.15}
\caption{\textbf{Main evaluation on BenchCAD-QA (\texttt{qa\_code} modality), by capability axis.} Per-model accuracy on the code-conditioned QA bank across the same four capability axes as Table~\ref{tab:main_eval_qa_img}. The two tables share questions and capability-axis definitions; only the conditioning modality differs (rendered images vs.\ source CadQuery). Score differences across the matched pair isolate the cost of replacing direct code access with visual recognition --- the \textbf{Holistic Spatial and Detailing Deficit}.}
\label{tab:main_eval_qa_code}
\begin{tabular}{l c c c c c}
\toprule
Model
 & \shortstack{Cadquery Code\\Recognition $\uparrow$ }
 & \shortstack{CAD Operation\\Understanding $\uparrow$ }
 & \shortstack{Industrial Parametric\\Abstraction $\uparrow$ }
 & \shortstack{Spatial\\Reasoning $\uparrow$ }
 & \shortstack{Total\\score $\uparrow$ } \\
\midrule
opus-4.7         & 0.868 & 0.800 & 0.793 & 0.595 & 0.801 \\
opus-4.7-thinking            & 0.891 & 0.781 & 0.851 & 0.632 & 0.829 \\
gemini-3.1-pro  & \textbf{0.914} & 0.782 & 0.867 & 0.537 & 0.836 \\
gemini-3.1-pro-thinking     & 0.907 & 0.783 & \textbf{0.876} & 0.537 & \textbf{0.838} \\
gpt-4o                              & 0.865 & 0.593 & 0.732 & 0.688 & 0.726 \\
gpt-5.3-chat-latest                 & 0.879 & \textbf{0.805} & 0.815 & 0.731 & 0.823 \\
gpt-5.3-thinking                    & 0.885 & 0.802 & 0.811 & 0.730 & 0.821 \\
moonshot-v1-128k-v     & 0.842    & 0.551    & 0.692    & \textbf{0.792}    & 0.700 \\                                                                       
  moonshot-v1-8k-v       & 0.772    & 0.603    & 0.555    & 0.536    & 0.620 \\               
openai-o3                & 0.804    & 0.701    & 0.689    & 0.492    & 0.708 \\                                                                
  google-gemma-4-31b-it    & 0.791    & 0.674    & 0.606    & 0.528    & 0.664 \\                                                                
  openai-gpt-oss-120b      & 0.790    & 0.732    & 0.656    & 0.379    & 0.689 \\
  nvidia-nemotron-3-120b   & 0.771    & 0.660    & 0.661    & 0.293    & 0.671 \\       
\midrule
\textit{blank code (baseline)}     & 0.040    & 0.257    & 0.290    & 0.420    & 0.223    \\
\bottomrule
\end{tabular}
\end{table*}

\begin{table*}[t]                           
  \centering                                                                                                                                          
  \small                                                          
  \setlength{\tabcolsep}{5pt}                                                                                                                         
  \renewcommand{\arraystretch}{1.15}                                                                                                                  
  \caption{\textbf{Vision2Code unified leaderboard on \texttt{BenchCAD}.} Specialist CAD lineage, frontier MLLMs, and our open Qwen3-VL-2B baseline.
   Best per block in \textbf{bold}.}                                                                                                   \label{tab:codegen_unified}                                                         
  \begin{tabular}{l c c c c c}                                                                                                                        
  \toprule                                                                                                                                            
  Model                                                                                                                                               
   & IoU $\uparrow$                                                                                                                                   
   & CD $\downarrow$                                                                                                                                  
   & ess\_score $\uparrow$
   & exec\_pct $\uparrow$                                                                                                                             
   & total $\uparrow$ \\                                          
  \midrule                                
  \multicolumn{6}{l}{\textit{Specialist CAD lineage ($\sim$2B)}} \\
  Cadrille-RL~\citep{Kolodiazhnyi2026}    & 0.0683          & 0.3507          & 0.3118          & 91.2          & 0.3382 \\                           
  CADEvolve v3~\citep{Elistratov2026}     & \textbf{0.7497} & \textbf{0.0080} & \textbf{0.3715} & \textbf{92.7} & \textbf{0.6014} \\
  \midrule                                                                                                                                            
  \multicolumn{6}{l}{\textit{Frontier MLLMs}} \\                                                                                                      
  gpt-4o                              & 0.1884          & 0.0623          & 0.4680          & 87.0          & 0.2274 \\                               
  gpt-5.3 (no thinking)               & 0.1952          & 0.0594          & 0.4940          & 69.5          & 0.2160 \\                               
  gpt-5.3 (thinking)                  & 0.2072          & 0.0566          & 0.4550          & 67.5          & 0.2120 \\                               
  claude-sonnet-4.6 (no thinking)     & 0.2380          & 0.0320          & 0.5060          & 85.4          & 0.3510 \\                               
  claude-sonnet-4.6 (thinking-high)   & 0.2420          & 0.0220          & --              & 90.0          & 0.3850 \\                               
  claude-opus-4.7 (no thinking)       & 0.2740          & 0.0210          & 0.4770          & \textbf{94.0} & 0.3830 \\                               
  claude-opus-4.7 (thinking)          & 0.2670          & 0.0240          & 0.4710          & 90.4          & 0.3780 \\                               
  gemini-3.1-pro (no thinking)        & 0.2560          & 0.0400          & 0.3780          & 74.0          & 0.3290 \\                               
  gemini-3.1-pro (thinking)           & \textbf{0.2790} & 0.0250          & \textbf{0.5670} & 79.8          & \textbf{0.3970} \\                      
  openai/o3                           & 0.5004          & 0.0143          & 0.4440          & 5.6           & 0.1081 \\                               
  moonshot-v1-128k                    & 0.1530          & 0.0572          & 0.1922          & 10.5          & 0.0612 \\                               
  moonshot-v1-8k                      & 0.2421          & 0.0577          & 0.1792          & 9.0           & 0.0604 \\                               
  \midrule                                                                                                                                            
  \multicolumn{6}{l}{\textit{Ours --- Qwen3-VL-2B}} \\                                                                                                
  qwen3-2b-sft-ood                    & 0.5630          & 0.0174          & 0.7380          & 94.2          & 0.5877 \\                               
  qwen3-2b-sft-iid                    & 0.6400          & 0.0108          & 0.8710          & 88.1          & 0.6282 \\                               
  qwen3-2b-rl-ood                     & 0.7140          & 0.0047          & 0.8100          & \textbf{99.1} & 0.7231 \\                               
  qwen3-2b-rl-iid                     & \textbf{0.7520} & \textbf{0.0041} & \textbf{0.8920} & 98.9          & \textbf{0.7682} \\                      
  qwen3-2b (baseline)                 & 0.0032          & 0.3652          & 0.0042          & 14.6          & 0.0084 \\                               
  \midrule                                                                                                                                            
  gpt-4o (black img)                  & 0.0802          & 0.1011          & 0.2244          & 87.0          & \textbf{0.1102} \\                      
  \bottomrule                                                                                                                                         
  \end{tabular}                                                                                                                                       
  \end{table*}          

% \end{table}

Table~\ref{tab:op_recall_train} reports the recall of operations for training models.
\begin{table*}[t]
  \centering
  \small
  \setlength{\tabcolsep}{5pt}
  \renewcommand{\arraystretch}{1.15}
  \caption{\textbf{Per-operation recall after BenchCAD training.}
  We compare the pretrained baseline, OOD-family SFT, and IID SFT on Vision2Code operation recall. BenchCAD training substantially improves recall on BenchCAD-distinguishing operations, especially \texttt{revolve}, \texttt{fillet}, \texttt{loft}, \texttt{sweep}, and array-based operations.}
  \label{tab:op_recall_train}
  \begin{tabular}{l c c c c}
  \toprule
  Operation & baseline & ood-sft & iid-sft & $|\text{GT}|$ \\
  \midrule
  \multicolumn{5}{l}{\emph{Basic ops (shared with prior corpora: cad-recode / text2cad)}} \\
  \texttt{extrude}   & \textbf{100.0} & 91.7           & 95.8           & 48  \\
  \texttt{cut}       & 87.2           & 70.2           & \textbf{89.4}  & 94  \\
  \texttt{union}     & \textbf{91.7}  & 41.7           & 79.2           & 48  \\
  \texttt{circle}    & 82.8           & 79.3           & \textbf{96.6}  & 58  \\
  \texttt{cylinder}  & 5.9            & 88.2           & \textbf{94.1}  & 34  \\
  \texttt{box}       & 2.6            & 84.6           & \textbf{92.3}  & 78  \\
  \texttt{rect}      & 5.0            & 65.0           & \textbf{90.0}  & 40  \\
  \texttt{workplane} & 0.0            & 77.4           & \textbf{88.7}  & 106 \\
  \midrule
  \multicolumn{5}{l}{\emph{Advanced ops (BenchCAD-distinguishing)}} \\
  \texttt{hole}         & 0.0            & 82.8           & \textbf{93.1}  & 58 \\
  \texttt{chamfer}      & 0.0            & \textbf{85.3}  & \textbf{85.3}  & 68 \\
  \texttt{revolve}      & 0.0            & 54.8           & \textbf{93.5}  & 62 \\
  \texttt{fillet}       & 0.0            & 25.0           & \textbf{87.5}  & 16 \\
  \texttt{loft}         & 0.0            & 22.2           & \textbf{77.8}  & 18 \\
  \texttt{rarray}       & 0.0            & 77.8           & \textbf{100.0} & 18 \\
  \texttt{sweep}        & 0.0            & 25.0           & \textbf{75.0}  & 16 \\
  \texttt{polarArray}   & 0.0            & 33.3           & \textbf{83.3}  & 12 \\
  \texttt{shell}        & 0.0            & 80.0           & \textbf{100.0} & 10 \\
  \texttt{sweep+helix}  & 0.0            & \textbf{100.0} & 50.0           & 4  \\
  \texttt{twistExtrude} & 0.0            & \textbf{100.0} & \textbf{100.0} & 2  \\
  \midrule
  \textbf{Macro recall} (16 ops) & 17.5\% & 59.9\% & \textbf{84.0\%} & --- \\
  \textbf{Exec rate}             & 90.0\% & 94.0\% & \textbf{99.0\%} & --- \\
  \bottomrule
  \end{tabular}
\end{table*}

\begin{table*}[t]                            
  \centering                                                                                                                      
  \small
  \setlength{\tabcolsep}{5pt}                 
  \renewcommand{\arraystretch}{1.15}      
  \caption{\textbf{The Family Cliff.} \texttt{img2cq} pass rate (\%) split by family difficulty tier. Even the strongest reasoning model collapses on
  the hard tier, exhibiting a $>$45-point gap from easy. The 26 hard-tier families exercise advanced operations (helical sweeps, twist-extrusion,     
  lofted Booleans) and parametric constraints (gear module--tooth--pitch consistency, spring free-length--turn-count coupling) that no evaluated model
   reliably preserves.}                                                                                                                               
  \label{tab:family_cliff}                                                                                                                            
  \begin{tabular}{l c c c c}
  \toprule                                                                                                                                            
  Model & Easy & Medium & Hard & $\Delta$ E--H \\                 
  \midrule
  claude-opus-4.7        & 57.1          & \textbf{62.5} & \textbf{45.5} & +11.7 \\                                                                   
  gpt-4o                 & 38.8          & 47.6          & 40.4          & $-$1.5 \\
  gpt-5.3 (thinking)     & 27.4          & 35.7          & 30.9          & $-$3.5 \\                                                                  
  gpt-5.3 (no thinking)  & 29.9          & 30.1          & 27.0          & +2.8  \\
  gemini-2.5-pro         & 22.0          & 28.1          & 20.2          & +1.8  \\                                                                   
  gpt-5.2                & 11.5          & 26.3          & 18.6          & $-$7.1 \\
  qwen3-VL-32B-Instruct  & 35.3          & 18.8          & 0.0           & +35.3 \\                                                                   
  Moonshot-v1-32k-vp     & 9.5           & 13.1          & 7.8           & +1.7  \\                                                                   
  Moonshot-v1-8k-vp      & 4.8           & 7.1           & 2.9           & +1.9  \\                                                                   
  Moonshot-v1-128k-vp    & 0.0           & 8.5           & 3.8           & $-$3.8 \\                                                                  
  \bottomrule                                                                                                                                         
  \end{tabular}                                                                                                                                       
  \end{table*}                 

% \begin{table}[t]
% \centering
% \small
% \caption{The Family Cliff.}
% \texttt{img2cq} pass rate (\%) split by family difficulty tier. Even the strongest reasoning model collapses on the hard tier, exhibiting a $>$45-point gap from easy. The 26 hard-tier families exercise advanced operations (helical sweeps, twist-extrusion, lofted Booleans) and parametric constraints (gear module--tooth--pitch consistency, spring free-length--turn-count coupling) that no evaluated model reliably preserves. \textcolor{red}{Numbers are placeholders.}
% \label{tab:family_cliff}
% \begin{tabular}{lcccc}
% \toprule
% Model & Easy & Medium & Hard & $\Delta$ E--H \\
% \midrule
% claude-opus-4.7 & 57.1 & 62.5 & 45.5 & +11.7 \\
% gpt-4o & 38.8 & 47.6 & 40.4 & -1.5 \\
% gpt-5.3 thinking & 27.4 & 35.7 & 30.9 & -3.5 \\
% gpt-5.3 (no thinking) & 29.9 & 30.1 & 27.0 & +2.8 \\
% gemini-2.5-pro & 22.0 & 28.1 & 20.2 & +1.8 \\
% gpt-5.2 & 11.5 & 26.3 & 18.6 & -7.1 \\
% qwen3-VL-32B-Instruct & 35.3 & 18.8 & 0.0 & +35.3 \\
% Moonshot-v1-32k-vp & 9.5 & 13.1 & 7.8 & +1.7 \\
% Moonshot-v1-8k-vp & 4.8 & 7.1 & 2.9 & +1.9 \\
% Moonshot-v1-128k-vp & 0.0 & 8.5 & 3.8 & -3.8 \\
% \bottomrule
% \end{tabular}
% \end{table}

\section{CAD Operational Blindspot: Per-Operation Recall}
\label{app:op_gap_full}

Table~\ref{tab:op_recall} reports operation-level recall on Vision2Code.
While both models recover common sketch-and-extrude patterns such as \texttt{extrude}, \texttt{circle}, \texttt{rect}, and \texttt{workplane}, recall remains much lower for industrially important operations such as \texttt{chamfer}, \texttt{revolve}, \texttt{threePointArc}, \texttt{shell}, and counterbored holes.
This gap suggests that models often approximate the final geometry with simpler primitives instead of recovering the intended parametric construction.
Thinking mode improves several planning-heavy operations, including \texttt{loft}, \texttt{polygon}, \texttt{fillet}, and \texttt{rarray}, but degrades common boolean and scaffolding operations such as \texttt{cut}, \texttt{cutBlind}, \texttt{workplane}, \texttt{sweep}, \texttt{box}, and \texttt{union}.
The resulting drop in macro recall and execution rate indicates that inference-time reasoning introduces a synthesis--execution trade-off rather than a uniform improvement.

\begin{table}[t]
    \centering
    \small
    \setlength{\tabcolsep}{4pt}
    \caption{\textbf{Per-operation recall on Vision2Code.}
    We compare gpt-5.3 with gpt-5.3-thinking on operation-level recall and execution rate.
    Thinking mode slightly lowers overall recall and executability, despite improving a few planning-heavy operations such as \texttt{loft}, \texttt{polygon}, and \texttt{fillet}.}
    \label{tab:op_recall}

  \begin{tabular}{lrrr}                                             
  \toprule                                                                                 
  Operation & gpt-5.3 & gpt-5.3-thk & $|\text{GT}|$ \\                                                             
  \midrule                                                                                                             
  \multicolumn{4}{l}{\emph{Basic ops (shared with prior corpora: cad-recode / text2cad)}} \\                           
  \texttt{extrude}              & 83.3          & \textbf{84.4} & 90  \\                                               
  \texttt{cut}                  & \textbf{44.1} & 32.2          & 59  \\                                               
  \texttt{union}                & \textbf{65.3} & 59.7          & 72  \\                                               
  \texttt{circle}               & \textbf{78.7} & 77.3          & 75  \\                                               
  \texttt{cylinder}             & 4.5           & \textbf{7.6}  & 66  \\                                               
  \texttt{box}                  & \textbf{42.3} & 34.6          & 78  \\                                               
  \texttt{rect}                 & 63.0          & \textbf{65.2} & 46  \\                                               
  \texttt{workplane}            & \textbf{86.3} & 77.4          & 124 \\                                               
  \midrule                                                                                                             
  \multicolumn{4}{l}{\emph{Advanced ops (BenchCAD-distinguishing)}} \\                                                 
  \texttt{hole}                 & \textbf{60.2} & 55.4          & 83  \\                                               
  \texttt{chamfer}              & \textbf{8.6}  & 4.3           & 70  \\                                               
  \texttt{cutThruAll}           & \textbf{16.2} & 10.8          & 37  \\                                               
  \texttt{revolve}              & \textbf{14.3} & 10.7          & 28  \\                                               
  \texttt{fillet}               & 0.0           & \textbf{12.0} & 25  \\
  \texttt{cutBlind}             & \textbf{36.8} & 26.3          & 19  \\                                               
  \textbf{\texttt{loft}}        & 5.9           & \textbf{23.5} & 17  \\
  \texttt{threePointArc}        & \textbf{6.2}  & 0.0           & 16  \\                                               
  \texttt{rarray}               & 7.1           & \textbf{14.3} & 14  \\                                               
  \textbf{\texttt{sweep}}       & \textbf{53.8} & 46.2          & 13  \\                                               
  \texttt{polygon}              & 54.5          & \textbf{72.7} & 11  \\                                               
  \texttt{makeHelix}            & 12.5          & 12.5          & 8   \\                                               
  \texttt{slot2D}               & 25.0          & 25.0          & 8   \\                                               
  \texttt{polarArray}           & 20.0          & 20.0          & 5   \\                                               
  \texttt{shell}                & \textbf{25.0} & 0.0           & 4   \\
  \texttt{mirrorY}              & 0.0           & 0.0           & 4   \\                                               
  \texttt{sphere}               & 100.0         & 100.0         & 4   \\
  \texttt{cboreHole}            & 0.0           & 0.0           & 3   \\                                               
  \texttt{radiusArc}            & \textbf{33.3} & 0.0           & 3   \\
  \midrule                                                                                                             
  \textbf{Macro recall} (advanced, 19 ops) & \textbf{25.2\%} & 22.8\% & --- \\
  \textbf{Macro recall} (all, 27 ops)      & \textbf{35.1\%} & 32.3\% & --- \\                                         
  \textbf{Exec rate}                       & \textbf{69.5\%} & 67.5\% & --- \\                                         
  \bottomrule                                                                                                          
  \end{tabular} 
  
\end{table}

\section{Datasheet for BenchCAD}
\label{app:datasheet}

We follow the datasheet template of Gebru et al.\ (2021); responses below are condensed for the appendix and reproduced in full as a separate Croissant 1.0 metadata file shipped with the dataset release.

\paragraph{Motivation.}
\textbf{For what purpose was the dataset created?} BenchCAD was created to evaluate the parametric-CAD capabilities of large vision-language and code-language models along three sub-capabilities (visual perception, parametric abstraction, code synthesis) on industrially representative part families with execution-verified ground truth.
\textbf{Who created the dataset and on behalf of which entity?} The authors. 
\textbf{Who funded the dataset?} The authors' individual funds. 

\paragraph{Composition.}
\textbf{What do the instances represent?} Each instance is a parametric CAD part: an executable CadQuery Python program, the resulting STEP file, four canonical-view PNG renders, a parameter JSON, and an op-list JSON.
\textbf{How many instances are there in total?} 17{,}900 verified parts in BenchCAD, 2{,}400 paired QA questions in BenchCAD-QA (each evaluated under both visual and code conditioning, yielding 4{,}800 records), and 748 curated edit pairs in BenchCAD-Edit.
\textbf{Does the dataset contain all possible instances?} No --- BenchCAD samples a subset of all the industry CAD generation part familys, and sampled a subset of parameter space per family; the schema permits unbounded sampling under the documented constraints.
\textbf{Is there a label/target?} The CadQuery source code, STEP, parameters, and op list jointly serve as ground truth; for QA tasks, gold answers are derived deterministically from parameters; for edit tasks, target code and target STEP are paired with the original.

\paragraph{Collection process.}
\textbf{How was the data acquired?} Procedurally generated by the BenchCAD generation pipeline (\S\ref{sec:dataset}); the only human-in-the-loop component is edit-pair curation by the authors.

\textbf{Who was involved in data collection?} The authors.
\textbf{Were any ethical review processes conducted?} Not applicable; no human-subjects data.

\paragraph{Preprocessing/cleaning/labeling.}
\textbf{Was preprocessing/cleaning done?} Verification (sandbox execution + non-degenerate volume + 30\,s timeout, followed by domain-expert visual sign-off) is the sole filter applied to BenchCAD.
The Fusion360 and DeepCAD subsets undergo additional curation: parsing failures and parts whose reconstruction yields invalid geometry are excluded.
\textbf{Was the raw data saved?} Yes; the unverified pre-filter output is retained for analysis and is available on request.

\paragraph{Uses.}
\textbf{Has the dataset been used for any tasks already?} The five tasks defined in \S\ref{sec:tasks}.
\textbf{Is there anything that prevents responsible reuse?} The dataset contains no PII, no copyrighted source designs, and no safety-sensitive specifications.
The standard-anchored families reference public ISO/DIN/EN/ASME/IEC standard codes by number and name; we do not redistribute the standard documents themselves.

\paragraph{Distribution.}
\textbf{Will the dataset be distributed?} Yes, on Hugging Face under \texttt{BenchCAD/BenchCAD}.
\textbf{When?} Released with this preprint.
\textbf{Under what licence?} CC-BY-4.0 for data; MIT for code (evaluation harness).
\textbf{Are there restrictions?} Standard CC-BY-4.0 attribution.

\paragraph{Maintenance.}
\textbf{Who will support the dataset?} The authors via the GitHub issue tracker and Hugging Face dataset card.
\textbf{How will errata be communicated?} Versioned releases on Hugging Face with changelogs in the dataset card.

\section{Question Bank Construction}
\label{app:qbank}

The numeric-QA tasks (\texttt{qa\_img}, \texttt{qa\_code}) draw from a per-family question bank constructed in three steps. (1) For each family, a domain-knowledge author writes 6--12 question \emph{templates} parameterised by the family's exposed parameters --- e.g.\ for \texttt{involute\_gear}: \emph{``what is the ratio of root-to-tip diameter?''}, \emph{``how many teeth does the visible gear have?''}.
Templates are typed (ratio, integer count, ordinal) and constrained to be scale-invariant. (2) Templates are instantiated per-record by deterministic substitution of the verified parameter values, yielding a gold answer alongside each question. (3) A second author reviews each instantiation for ambiguity and visibility (a question must be answerable from the four canonical views with no occluded features).
The final bank contains 2{,}400 unique (question, gold answer) pairs sampled across 17{,}900 records --- each evaluated under both visual (\texttt{qa\_img}) and code (\texttt{qa\_code}) conditioning, yielding 4{,}800 total QA records.
The composition is approximately 60\% ratio, 35\% integer count, and 5\% ordinal. We release the question-template source code so that the bank can be regenerated under alternative parameter samples.

\section{QA System Prompts}
\label{app:qa-system-prompts}

To ensure reproducibility, we provide the exact system prompts used in our QA evaluation.
All models are evaluated with the same prompt template. The model is required to output
only a JSON array of numbers, which enables automatic numeric grading. For binary
questions, we encode ``yes'' as 1 and ``no'' as 0.

\subsection{Image-based QA System Prompt}
\label{app:qa-img-system-prompt}

\begin{tcolorbox}[
    title=\texttt{QA\_IMG\_SYSTEM\_PROMPT},
    colback=gray!4,
    colframe=gray!45,
    fonttitle=\bfseries,
    breakable,
    sharp corners,
    boxrule=0.5pt,
]
\small
You are an expert CAD engineer. You will be shown a
\texttt{DIAGONAL\_VIEW\_LAYOUT}. You will be given a list of numeric
questions about the part. Answer each with a single number.

\vspace{0.5em}
\textbf{Rules:}
\begin{itemize}
    \item Output ONLY a JSON array of numbers, one per question, in the same order.
    \item No text, no keys, no explanation. Just the array.
    \item For yes/no questions, use 1 for yes and 0 for no.
    \item For count questions, use an integer, e.g., 12, not ``twelve''.
    \item For ratio questions, use a decimal, e.g., 2.5.
    \item For dimensional questions, answer in whatever consistent unit the code uses;
    the scale is arbitrary and the grader compares numeric magnitude.
\end{itemize}

\textbf{Example input:}
\begin{verbatim}
["How many teeth?", "What is the module?"]
\end{verbatim}

\textbf{Example output:}
\begin{verbatim}
[20, 2.5]
\end{verbatim}
\end{tcolorbox}

\subsection{Code-based QA System Prompt}
\label{app:qa-code-system-prompt}

\begin{tcolorbox}[
    title=\texttt{QA\_CODE\_SYSTEM\_PROMPT},
    colback=gray!4,
    colframe=gray!45,
    fonttitle=\bfseries,
    breakable,
    sharp corners,
    boxrule=0.5pt,
]

\small
You are an expert CAD engineer. You will be shown CadQuery Python code for a
mechanical part. You will be given a list of numeric questions about the part
this code produces.

\vspace{0.5em}
\textbf{Rules:}
\begin{itemize}
    \item Output ONLY a JSON array of numbers, one per question, in the same order.
    \item No text, no keys, no explanation. Just the array.
    \item For yes/no questions, use 1 for yes and 0 for no.
    \item For count questions, use an integer, e.g., 12, not ``twelve''.
    \item For ratio questions, use a decimal, e.g., 2.5.
    \item For dimensional questions, answer using the same scale as the numeric
    literals in the code.
\end{itemize}

\textbf{Example input code:} The code creates a gear with 20 teeth and module 2.5.

\textbf{Example input questions:}
\begin{verbatim}
["How many teeth in the gear?", "What is the module?"]
\end{verbatim}

\textbf{Example output:}
\begin{verbatim}
[20, 2.5]
\end{verbatim}
\end{tcolorbox}

\section{Code Generation System Prompts}
\label{app:codegen-system-prompts}

To ensure reproducibility, we provide the exact prompts used for code generation.
The model is given a normalized four-view composite render of an industrial part and
is asked to generate executable CadQuery Python code. All generated code is evaluated
under the same execution protocol: \texttt{cadquery} is pre-imported as \texttt{cq},
and the final reconstructed solid must be stored in \texttt{result}.

\subsection{Primary Vision-to-CAD System Prompt}
\label{app:primary-vision-to-cad-system-prompt}

\begin{tcolorbox}[
    title=\texttt{SYSTEM\_PROMPT},
    colback=gray!4,
    colframe=gray!45,
    fonttitle=\bfseries,
    breakable,
    sharp corners,
    boxrule=0.5pt
]
\small
Generate CadQuery Python code from a \texttt{DIAGONAL\_VIEW\_LAYOUT}.

\vspace{0.5em}
Renders are normalized: bbox centered at \texttt{[0.5,0.5,0.5]}, longest side maps to \texttt{[0,1]}.
Match the orientation exactly---do not rotate or remap axes. World XYZ in your code must match world XYZ in the renders.

\begin{itemize}
    \item \texttt{cadquery} is pre-imported as \texttt{cq}; no imports, no \texttt{show\_object}.
    \item Store final solid in \texttt{result}.
    \item Output ONLY code, no explanation or markdown.
\end{itemize}
\end{tcolorbox}

\subsection{Vision-to-CAD User Prompt}
\label{app:vision-to-cad-user-prompt}

\begin{tcolorbox}[
    title=\texttt{USER\_PROMPT},
    colback=gray!4,
    colframe=gray!45,
    fonttitle=\bfseries,
    breakable,
    sharp corners,
    boxrule=0.5pt
]
\small
Generate CadQuery code to recreate this industrial part shown in the 4-view composite render.
\end{tcolorbox}

\subsection{Cadrille Baseline System Prompt}
\label{app:cadrille-system-prompt}

\begin{tcolorbox}[
    title=\texttt{CADRILLE\_SYSTEM\_PROMPT},
    colback=gray!4,
    colframe=gray!45,
    fonttitle=\bfseries,
    breakable,
    sharp corners,
    boxrule=0.5pt
]
\small
You are a CadQuery expert. Given a normalized \texttt{DIAGONAL\_VIEW\_LAYOUT}.

\vspace{0.5em}
Write CadQuery Python code that reproduces the geometry. Output ONLY Python code.
\end{tcolorbox}

\subsection{Edit-task System Prompt}
\label{app:edit-code-system-prompt}
\begin{tcolorbox}[title=\texttt{EDIT\_CODE\_SYSTEM\_PROMPT}, colback=gray!4, colframe=gray!45,fonttitle=\bfseries, breakable, sharp corners, boxrule=0.5pt,]                                   
  \small                                                         
  You are an expert CAD engineer. You will be given:
  \begin{enumerate}                                                   \item A CadQuery Python script that builds a parametric mechanical part.
      \item A natural-language edit instruction describing a change to one specific feature.                
  \end{enumerate}
    
  Your task: return the script with the MINIMAL edit that achieves the requested feature change. Only the feature(s) the instruction explicitly mentions should change; every other feature, dimension, and detail must stay exactly as in the original.    
  \vspace{0.5em}                                                  \textbf{Rules:}
  \begin{itemize}
  \item Output ONLY executable Python code, no explanation, no markdown fences.
  \item Make the smallest set of changes that realize the instruction. Do NOT touch unrelated literals, operations, or structure.
  \item Do NOT refactor, rename variables, reorder operations, or add/remove imports.
  \item If the instruction adds a new feature: append the operation in the most natural place; keep the rest of the chain intact.
  \item If the instruction removes a feature: delete just that operation block; do not rewrite surrounding code.
  \item If the instruction modifies a dimension/parameter: update only the literal(s) that depend on it (and, if present, the matching parameter comment).
  \end{itemize}
  \end{tcolorbox}                                                 
  
\label{app:edit-img-gt-system-prompt}                             \begin{tcolorbox}[
title=\texttt{EDIT\_IMG\_GT\_SYSTEM\_PROMPT},               
colback=gray!4, colframe=gray!45, fonttitle=\bfseries, breakable, sharp corners, boxrule=0.5pt,]
  \small
  You are an expert CAD engineer. You will be given:              
  \begin{enumerate}
  \item A CadQuery Python script that builds a parametric mechanical part(CURRENT state).
  \item A 2$\times$2 composite image of the TARGET geometry from 4 diagonal viewpoints.                                         
  \end{enumerate}                                                
  Your task: return the script with the MINIMAL edit that transforms the CURRENT geometry (described by the code) into the TARGET geometry (shown in the image). The image is the only specification of the desired change, there is NO text instruction. Compare the code's geometry against the image and infer which single feature differs.                      
  Only the feature(s) that differ between code and image should change; every other feature, dimension, and detail must stay exactly as in the original code.
    
  \vspace{0.5em}
  \textbf{Rules:}                                                               
  \begin{itemize}
  \item Output ONLY executable Python code, no explanation, no markdown fences.
  \item Make the smallest set of changes that realize the visual difference. Do NOT touch unrelated literals, operations, or structure.
  \item Do NOT refactor, rename variables, reorder operations, or add/remove imports.
  \item If the target image shows a new feature that is absent in the code: append the operation in the most natural place; keep the rest of the chain intact.
  \item If the target image shows a feature removed: delete just that operation block; do not rewrite surrounding code.
  \item If a dimension/parameter has changed: update only the literal(s) that control that dimension (and the matching parameter comment if present).  
  \end{itemize}
  \end{tcolorbox}

\label{app:edit-img-system-prompt}

\begin{tcolorbox}[
    title=\texttt{EDIT\_CODE\_IMG\_SYSTEM\_PROMPT (Ablation)},
    colback=gray!4,
    colframe=gray!45,
    fonttitle=\bfseries,
    breakable,
    sharp corners,
    boxrule=0.5pt,
]
\small
You are an expert CAD engineer. You will be given:
\begin{enumerate}
    \item A CadQuery Python script that builds a parametric mechanical part.
    \item A natural-language edit instruction describing a change to one specific feature.
    \item A 2$\times$2 composite image of the part from 4 diagonal viewpoints (CURRENT state) as an additional visual reference for the script.
\end{enumerate}

Your task: return the script with the MINIMAL edit that achieves the
requested feature change. Only the feature(s) the instruction explicitly
mentions should change; every other feature, dimension, and detail must
stay exactly as in the original.

\vspace{0.5em}
\textbf{Rules:}
\begin{itemize}
    \item Output ONLY executable Python code, no explanation, no markdown fences.
    \item Make the smallest set of changes that realize the instruction. Do NOT touch unrelated literals, operations, or structure.
    \item Do NOT refactor, rename variables, reorder operations, or add/remove imports.
    \item If the instruction adds a new feature: append the operation in the most natural place; keep the rest of the chain intact.
    \item If the instruction removes a feature: delete just that operation block; do not rewrite surrounding code.
    \item If the instruction modifies a dimension/parameter: update only the literal(s) that depend on it (and, if present, the matching parameter comment).
\end{itemize}
\end{tcolorbox}

\section{Rotation-Invariant IoU}
\label{app:iou}

We compute IoU under a configurable rotation-invariant protocol, taking the maximum over either the 6-element face-up cube symmetry group or the full 24-element cube rotation group. Before voxelization, each part is normalized by centering it at the origin and scaling it by the largest semi-axis of its bounding box. For each candidate rotation $g \in G$, we rotate the predicted voxel grid $\hat{V}$ and compute its overlap with the reference voxel grid $V$; the reported score is
\[
\max_{g \in G} \mathrm{IoU}(g \cdot \hat{V}, V).
\]
This metric is used diagnostically rather than as the primary score: a large gain from standard IoU to 24-rotation IoU indicates that the model may have generated the right geometry on the wrong construction plane. This failure is partly induced by a workplane prior: in online database, and manufacturing-oriented modeling workflows, parts are commonly initialized on the \texttt{XY} plane as the default sketching or setup plane. As a result, models may over-prefer \texttt{XY}-based constructions even when the target geometry is defined on \texttt{XZ} or \texttt{YZ}. Consistent with this interpretation, Table~\ref{tab:rot24_base_plane} shows substantially larger rotation-IoU gains for \texttt{XZ}/\texttt{YZ} ground-truth parts than for \texttt{XY} parts.
\begin{table}[t]
\centering
\small
\setlength{\tabcolsep}{6pt}
\renewcommand{\arraystretch}{1.08}
\caption{\textbf{Effect of rotation-invariant IoU by GT base plane.}
For GPT-4o, we report $\Delta=\mathrm{IoU}_{\mathrm{rot24}}-\mathrm{IoU}_{\mathrm{single\text{-}axis}}$ over 164 valid 24-IoU samples with extrusion. Larger $\Delta$ indicates cases where the generated shape is geometrically similar but expressed in a mismatched coordinate frame.}
\label{tab:rot24_base_plane}
\begin{tabular}{@{}lrrrr@{}}
\toprule
GT base plane & $n$ & Mean $\Delta$ & Max $\Delta$ \\
\midrule
XY  & 70  & 0.0463 & 0.5907 \\
XZ  & 40  & 0.2305 & 0.8031 \\
YZ  & 54  & 0.2140 & 0.7053 \\
\bottomrule
\end{tabular}
\end{table}

\section{Edit Protocol}
\label{app:edit_protocol}

\paragraph{Pair construction.} The 748 BenchCAD-Edit pairs are drawn balanced across 106 families and four edit categories (balanced across dimensional, additive, subtractive, and multi-step categories).
For each pair, we (i) start from a verified BenchCAD record, (ii) apply a parameter or op-list edit yielding a target part, (iii) verify the target satisfies the same verification pipeline, (iv) write a natural-language instruction by template (\emph{``increase the bore diameter to 12\,mm''}) reviewed for unambiguity by a second author, and (v) cross-validate by running two strong models and inspecting any unexpected failure modes.
Pairs where models systematically disagree on instruction interpretation are revised or removed.

\paragraph{Scoring.} See Eq.~\ref{eq:edit_acc} for metric definitions (normalised IoU used in edit-task score). The non-target preservation check is computed on the spatial complement of the bounding region of the targeted feature (precomputed bounding-box annotation per pair).

\paragraph{Task taxonomy.} The five edit task types (T1--T5) are defined in Table~\ref{tab:task_taxonomy}. Difficulty rises monotonically with the structural complexity of the minimal correct diff, from a single literal swap (T1) to coordinated multi-block restructures (T5). The T1--T5 axis is orthogonal to the dimensional/additive/subtractive/multi-step categorisation used in the main paper: T1 and T3 are predominantly dimensional, T2 and T4 cover additive/subtractive, and T5 spans the multi-step regime.

\begin{table}[t]
    \centering
    \caption{\textbf{BenchCAD-Edit task taxonomy.} The 748 edit pairs are stratified into five task types T1--T5 by the structural form of the minimal correct diff. Difficulty rises monotonically from T1 (one-literal swap) to T5 (multi-block restructure with trig or coordinated sub-feature changes).}
    \label{tab:task_taxonomy}
    \small
    \begin{tabular}{@{}p{0.07\linewidth}p{0.18\linewidth}p{0.28\linewidth}p{0.18\linewidth}p{0.19\linewidth}@{}}
    \toprule
    \textbf{Code} & \textbf{Name} & \textbf{Definition} & \textbf{Diff signature} & \textbf{Example instruction} \\
    \midrule
    T1 \mbox{(n=180)} & literal\_\allowbreak replace & Replace a single numeric literal that the instruction names explicitly (\emph{from X to Y}). & one literal swap; equal lines added/removed & \emph{``Increase the cylinder radius from 4.5 to 5.85.''} \\
    T2 \mbox{(n=104)} & chain\_\allowbreak transform & Append a single chainable whole-part transform (rotate / mirror / array). & one \texttt{result = result.<op>(...)} line appended & \emph{``Rotate the entire part by 90 degrees about the X axis.''} \\
    T3 \mbox{(n=111)} & relative\_\allowbreak compute & Read an existing literal and derive another value (ratio, percentage, $2\times$, $1/4$). & one literal swap, value derived from another literal & \emph{``Set the central bore diameter equal to twice the hub-cylinder height.''} \\
    T4 \mbox{(n=199)} & feature\_\allowbreak edit & Add or delete a single CSG sub-block (hole, slot, recess, chamfer, fillet, boss). & one 3--7 line CSG block inserted or removed & \emph{``Drill a 6 diameter through-hole along the Z axis.''} \\
    T5 \mbox{(n=154)} & geometry\_\allowbreak rebuild & Non-trivial restructure: trig-derived slopes/chamfers/lofts, or paragraph-style rebuilds where two or more sub-features change in concert. & multi-block rewrite, trig in code, paragraph instruction & \emph{``Replace the ramped arms and raised center with a single flat cross-shaped solid of uniform thickness.''} \\
    \bottomrule
    \end{tabular}
    \end{table}

\begin{figure}[t]                       
    \centering
    \includegraphics[width=0.95\linewidth]{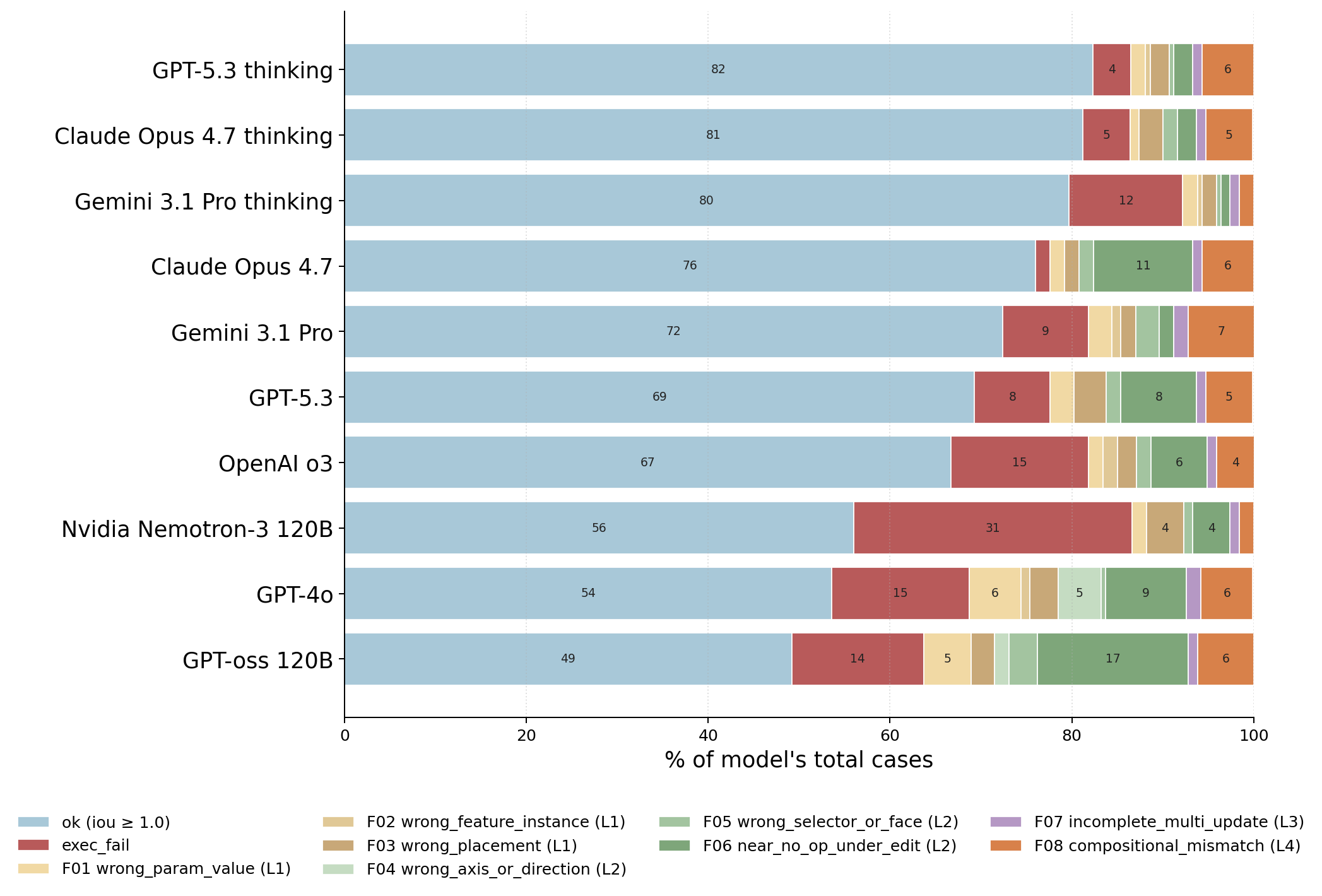}
    \caption{\textbf{Per-model failure-mode distribution on BenchCAD-Edit.} Each
   horizontal bar disaggregates one model's predictions into \texttt{ok},
  \texttt{exec\_fail}, and the eight semantic failure modes F01--F08 defined in 
  Table~\ref{tab:failure_codes}; segments are coloured by the underlying
  capability layer (sand $=$ L1 \emph{Holistic Visual Understanding}, sage $=$  
  L2 \emph{CAD Operations Comprehension}, purple $=$ L3 \emph{Industrial
  Parametric Abstraction}, orange $=$ L4 \emph{Spatial Reasoning $+$ Code       
  Synthesis}), so the L-mass within each bar reads off directly. Models are
  ordered by \texttt{ok} rate. The mass shifts systematically across model    
  generations: older or smaller-capacity systems concentrate failures in the L2
  band (F04--F06) and incur a non-trivial \texttt{exec\_fail} tail; recent large
   models without explicit reasoning largely close the L2 gap and leave residual
   mass on L1 (F01--F03); reasoning-tier closed models flatten L1 and L2 alike
  but expose an L3--L4 ceiling (F07--F08) on multi-instance and T5-style
  coordinated edits, indicating that thinking lifts capability up the
  L-hierarchy rather than uniformly reducing all error types.}
    \label{fig:edit_F_bars}
  \end{figure}

\paragraph{Failure taxonomy.} Every failing prediction is bucketed into one of eight semantic failure modes F01--F08 (Table~\ref{tab:failure_codes}, Fig~\ref{fig:edit_F_bars}), each mapped to one of the four capability layers L1--L4: F01--F03 isolate L1 \emph{Holistic Visual Understanding} (wrong value, wrong instance, wrong placement), F04--F06 isolate L2 \emph{CAD Operations Comprehension} (wrong axis, wrong selector, near no-op), F07 isolates L3 \emph{Industrial Parametric Abstraction} (incomplete multi-instance update), and F08 isolates L4 \emph{Spatial Reasoning $+$ Code Synthesis} (compositional geometry mismatch). Each label combines an automatic detection cue --- execution status, generated-vs-original/target IoU thresholds, and topology checks with a hand audit by the authors against the predicted code and rendered geometry; the rule set and L-mapping are fixed across all ten models reported, and \texttt{ok} (strict pass) and \texttt{exec\_fail} are tracked separately as the non-failure and gross-execution categories.

\begin{table}[t]
\centering
\caption{\textbf{Failure-mode taxonomy (F-codes).} Eight semantic failure modes used in the per-model failure analysis (Fig.~\ref{fig:edit_F_bars}). Each F-code is mapped to one capability layer: L1 \emph{Holistic Visual Understanding}, L2 \emph{CAD Operations Comprehension}, L3 \emph{Industrial Parametric Abstraction}, L4 \emph{Spatial Reasoning $+$ Code Synthesis}. Hand-labelled on the failing predictions of every BenchCAD-Edit run.}
\label{tab:failure_codes}
\small
\setlength{\tabcolsep}{4pt}
\renewcommand{\arraystretch}{1.15}
% allow line breaks at underscores in fixed-width name cells
\newcommand{\fname}[1]{\seqsplitfallback{#1}}
\providecommand{\seqsplitfallback}[1]{#1}
\begin{tabular}{@{}p{0.06\linewidth}p{0.18\linewidth}p{0.28\linewidth}p{0.18\linewidth}p{0.18\linewidth}@{}}
\toprule
\textbf{Code} & \textbf{Name} & \textbf{Definition} & \textbf{Detection cue} & \textbf{Representative failure} \\
\midrule
F01 \mbox{(L1)} & wrong\_\allowbreak param\_\allowbreak value & Right operation, wrong numerical value: dim, diameter, radius, length, or angle written off-target; geometry shape-correct but a single number misses. & $\mathrm{IoU}_{\text{gen}} \geq 0.9$, output differs from target by a single literal. & Model writes radius $5.6$ when prompt asks for $5.85$; cylinder topology unchanged. \\
F02 \mbox{(L1)} & wrong\_\allowbreak feature\_\allowbreak instance & Among multiple similar features (front vs.\ back ring, two legs, several circles), the model edits the wrong one or perturbs a neighbour. & $\mathrm{IoU}_{\text{gen}}$ swings markedly above OR below baseline; topology preserved. & Edit asks to enlarge the rear bore; model enlarges the front bore. \\
F03 \mbox{(L1)} & wrong\_\allowbreak placement & Right operation and value, but feature placed at slightly wrong coordinate, face, or orientation. & $\mathrm{IoU}_{\text{gen}} \geq 0.95$, small rigid offset against target. & Hole drilled $2$\,mm off-centre on a face whose origin the model misread. \\
F04 \mbox{(L2)} & wrong\_\allowbreak axis\_\allowbreak or\_\allowbreak direction & Rotation, mirror, or extrude direction wrong: rotate axis tuple permuted, mirror plane misidentified, sweep tangent flipped. & \texttt{status==ok}; rotated/mirrored bbox mismatches target axis. & gpt-4o rotate: \texttt{rotate((axis), (origin), deg)} with first two args swapped. \\
F05 \mbox{(L2)} & wrong\_\allowbreak selector\_\allowbreak or\_\allowbreak face & \texttt{.faces()}/\texttt{.edges()} selector returns wrong subset: \texttt{>Y} chosen on a Z-up part, OR-selector returns multi-face, lofted-face selector silently misses. & \texttt{status==ok}; operation applied to wrong topology vs.\ target. & Chamfer applied to all top edges instead of the single outer rim. \\
F06 \mbox{(L2)} & near\_\allowbreak no\_\allowbreak op\_\allowbreak under\_\allowbreak edit & Output $\approx$ original: comment-only edit, deleted wrong line, retained the feature meant to be removed, or applied the edit with no observable geometric effect. & $|\mathrm{IoU}_{\text{gen}} - \mathrm{IoU}_{\text{orig}}| < 0.005$. & claude-opus-4-7 (no-think): on $49\%$ of the failed records that rewrites the original byte-for-byte. \\
F07 \mbox{(L3)} & incomplete\_\allowbreak multi\_\allowbreak update & Edit requires updating multiple symmetric or repeated instances (helical-sweep points, replicated bolts/legs); model updates only a subset. & $\mathrm{IoU}_{\text{gen}}$ plateaus below $1.0$ across attempts; partial symmetry break. & Cross-parameter dim asks to scale all four legs; model scales only two. \\
F08 \mbox{(L4)} & compositional\_\allowbreak geometry\_\allowbreak mismatch & Composite or structural shape (slope strip, loft, sweep cross-section, multi-feature reorganisation) is partially right but does not match GT topology. & \texttt{status==ok}; $\mathrm{IoU}_{\text{gen}} \in [0.4, 0.95]$ on a T5-style edit. & Slope edit produces a chamfer of correct angle but along the wrong face strip. \\
\bottomrule
\end{tabular}
\end{table}

\begin{figure}[t]
  \centering
  \includegraphics[width=0.95\linewidth]{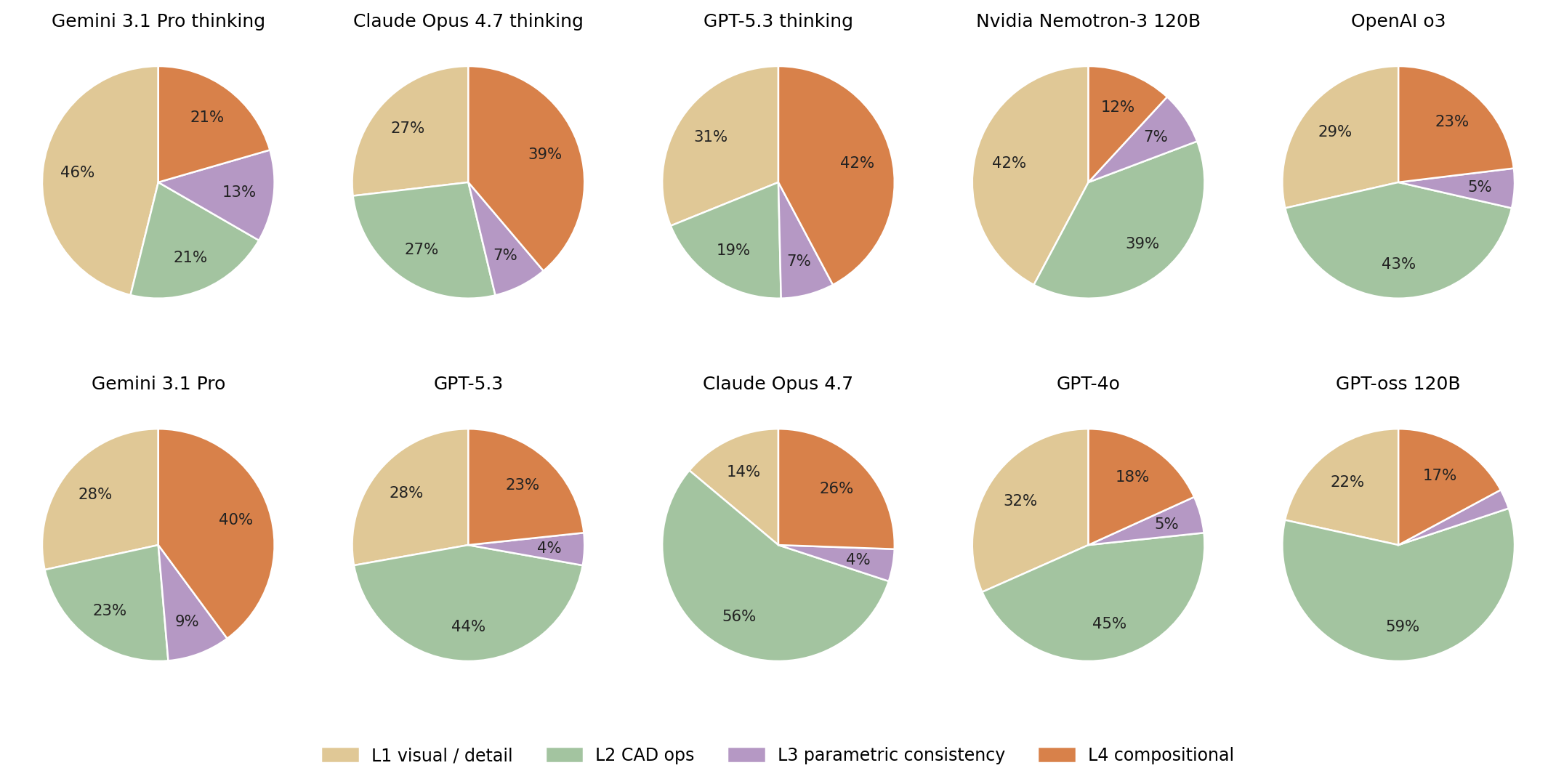}
  \caption{\textbf{Per-model failure-layer distribution on BenchCAD-Edit.} Each pie shows one model's failures aggregated by capability layer L1--L4 (collapsing the eight F-codes via the mapping in Table~\ref{tab:failure_codes}: F01--F03 $\to$ L1, F04--F06 $\to$ L2, F07 $\to$ L3, F08 $\to$ L4) and re-normalised so L1$+$L2$+$L3$+$L4 $= 100\%$ --- i.e.\ the pies show the \emph{shape} of each model's failure mode mix, independent of how often it fails overall; the title above each pie reports the absolute L-fail rate (\% of all predictions) so that scale is preserved. Models are arranged in order of increasing total L-fail rate (least-broken first). The mass shifts systematically along the L-axis across model generations: older or smaller-capacity systems concentrate failures in L2 (CAD-API misuse) and incur a non-trivial \texttt{exec\_fail} tail (counted separately, not shown); recent large models without explicit reasoning largely close the L2 gap and leave residual mass on L1 (visual/detail mistakes); reasoning-tier closed models flatten L1 and L2 alike but expose an L3--L4 ceiling on multi-instance and T5-style coordinated edits, indicating that thinking lifts capability up the L-hierarchy rather than uniformly reducing all error types.}
  \label{fig:edit_L_pies}

\end{figure}
\begin{figure}[t]
  \centering
  \includegraphics[width=0.95\linewidth]{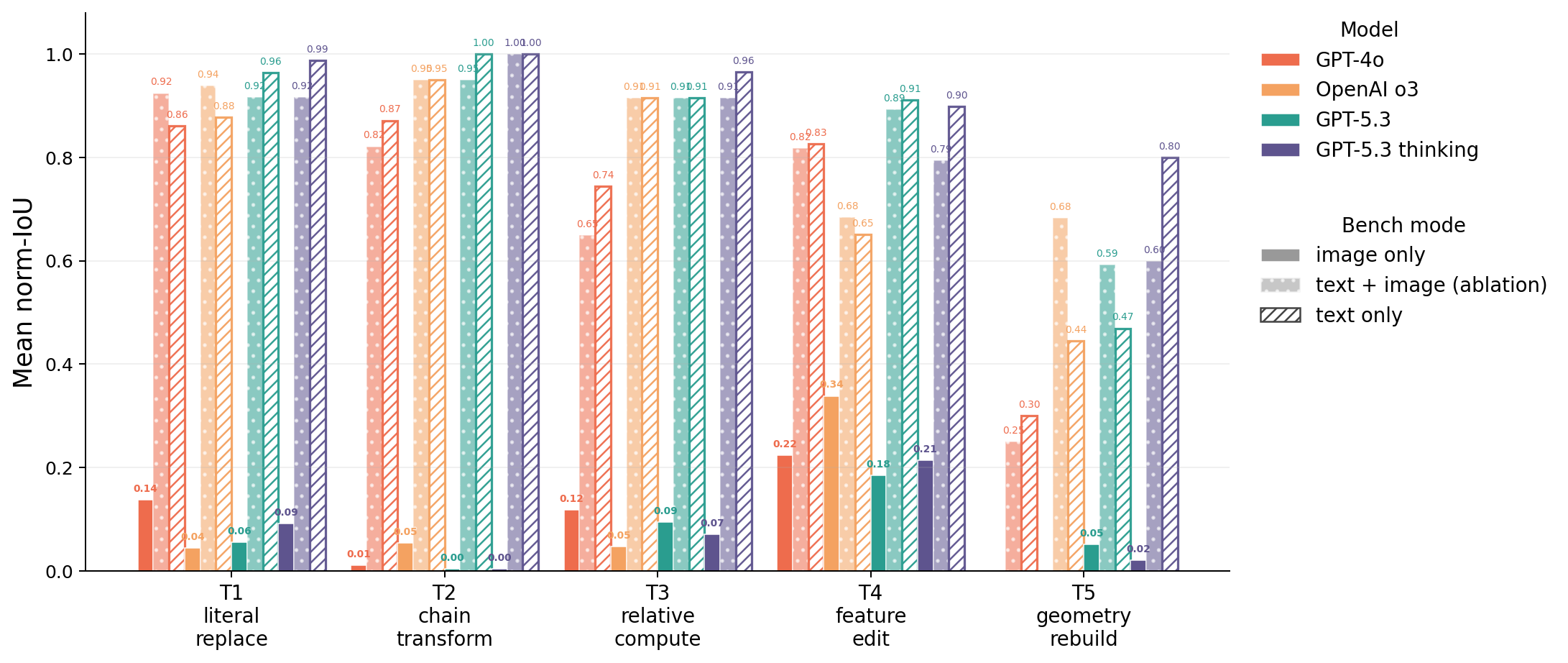}
  \caption{\textbf{BenchCAD-Edit under three input protocols, by
  task type.} Three protocols on the same $100$-pair subset, four OpenAI models. \emph{text}: original code $+$ NL instruction (main bench, \texttt{EDIT\_CODE\_SYSTEM\_PROMPT}). \emph{ablation}: original code $+$ NL instruction $+$ a four-view render of the original part (\texttt{EDIT\_IMG\_SYSTEM\_PROMPT}, App.~\ref{app:edit-img-system-prompt}). \emph{image-only}: original code $+$ four-view render of the target solid, no NL instruction (\texttt{EDIT\_IMG\_GT\_SYSTEM\_PROMPT}, App.~\ref{app:edit-img-gt-system-prompt}). Bars are \texttt{mean\_norm} (Eq.~\eqref{eq:edit_acc}) per task type. Four findings. (1) \emph{Adding the original image to a text instruction barely helps and sometimes hurts}: three of four models stay within $\pm 2$\,pt of text-only on the aggregate, indicating the NL instruction already supplies most of the signal and the visual reference adds little new information. (2) \emph{The one consistent winner is \texttt{o3}} ($+7$\,pt over text-only on aggregate, with the largest lift on T5), suggesting that for an instruction-following reasoning model the image disambiguates ambiguous referents that text alone leaves vague. (3) \emph{\texttt{gpt-5.3-thinking} is hurt by the image} ($-8$\,pt on T1 and $-20$\,pt on T5 vs.\ text-only); the thinking process appears to over-interpret the redundant visual signal, which is consistent with the same model being the worst image-only performer. (4) \emph{Text-only is the ceiling; image-only is the floor}: removing the NL instruction collapses every model to $0.04$--$0.34$ \texttt{mean\_norm} because a render shows what the part should look like but never tells the model that a radius is $5.85$, so dimensional edits (T1, T3) drop to near-zero, T5 floors at $0$, and only T2/T4 (add/remove a feature) retain weak signal. Net: the textual instruction does the heavy lifting and sets the upper bound; the original image is at best a clarifier (\texttt{o3}) and at worst a distractor for thinking models; image-only is a strict lower bound that primarily probes feature-presence reasoning rather than parametric exactness.}
  \label{fig:edit_image_type}
\end{figure}

\paragraph{Image-conditioned variant.} To probe whether the textual instruction itself is a confound, i.e.\ whether models succeed because the instruction states the change explicitly rather than because they reason about the geometry, we additionally evaluate an image-conditioned setting in which the natural-language instruction is replaced by a four-view render of the target solid (\texttt{EDIT\_IMG\_GT\_SYSTEM\_PROMPT}, App.~\ref{app:edit-img-gt-system-prompt}); the model receives the original code and must infer the edit visually. Across the four OpenAI models we tested, \texttt{mean\_norm} drops by $0.40$--$0.85$\,pt at every task type (Fig.~\ref{fig:edit_image_type}). The dominant failure mode is that a render specifies geometry but not numbers: the model can usually tell whether a feature is added or removed (T2/T4 retain weak signal), but cannot recover the exact dimension a textual ``from $X$ to $Y$'' would have stated, so dimensional edits (T1, T3) collapse to near-zero and trig-driven rebuilds (T5) floor uniformly at $0$. Image-only conditioning is therefore a strict lower bound on the text protocol and, in its current form, primarily isolates feature-presence reasoning rather than parametric exactness.

\paragraph{Does adding the image really help?} If image-only is a lower bound, the more interesting question is whether \emph{augmenting} the text instruction with the same four-view render of the original part (\texttt{EDIT\_IMG\_SYSTEM\_PROMPT}, App.~\ref{app:edit-img-system-prompt}) gives the model usable extra signal. We rerun the same four models in this third protocol and find the answer is mostly no. Three of the four models stay within $\pm 2$\,pt of text-only on the aggregate (Fig.~\ref{fig:edit_image_type}, ablation bars), so the NL instruction already supplies most of the signal and the visual reference adds little new information. The two exceptions are diagnostic: \texttt{o3} gains $\sim 7$\,pt on aggregate, with the largest lift on T5, suggesting that for an instruction-following reasoning model the image disambiguates ambiguous referents (\emph{the longer end}, \emph{the central pillar}) that text alone leaves vague; \texttt{gpt-5.3-thinking} \emph{loses} $7$\,pt on T1 and $20$\,pt on T5, mirroring its weak image-only performance and consistent with the thinking process over-interpreting a redundant visual signal. Net: the textual instruction does the heavy lifting and sets the upper bound; the original image is at best a clarifier and at worst a distractor for thinking models.

\paragraph{Supplementary results.} Figure~\ref{fig:edit_F_bars} reports failure attributes of all 748 eidt-pairs from 10 LLMs.
Figure~\ref{fig:edit_L_pies} disaggregates per-model failures into the L1--L4 buckets defined in Figure~\ref{fig:capability_hierarchy}. Figure~\ref{fig:edit_image_type} reports the parallel image-based setting (target render in lieu of an NL instruction) broken down by T1--T5.

\section{Scoring-Protocol Ablations}
\label{app:ablation}

\begin{table}[t]
\centering
\small
\setlength{\tabcolsep}{6pt}
\caption{\textbf{Scoring-protocol ablations.} (i) \emph{24-axial vs.\ single-axis IoU}: comparing the cube-symmetry-group IoU (max over 24 rotations) to the standard axis-aligned IoU isolates wrong-plane errors --- when 24-axial $>$ single-axis on the same prediction, the geometry is correct but emitted on the wrong base workplane (\S\ref{sec:exp_failure_analysis}). (ii) \emph{Single image vs.\ multi-view input}: collapsing the four canonical orthographic views into a single front view removes the multi-view evidence the model relies on for depth and back-side features. 
% \textcolor{red}{Numbers are placeholders pending the ablation run.}
}
\label{tab:ablation}
\begin{tabular}{lrr}
\toprule
Variant & total & $\Delta$ \\
\midrule
\multicolumn{3}{l}{\emph{IoU scoring protocol (Vision2Code, fixed model output)}} \\
24-axial rotation-invariant IoU        & 0.308   & 0.068 \\
single-axis IoU (axis-aligned only, default)               & 0.240   & ref \\
\midrule
\multicolumn{3}{l}{\emph{Visual input - view count (Vision2Code, single model)}} \\
4-view canonical orthographic (default)           & 0.240   & ref \\
single front view only                            & 0.215   & -0.025 \\
% \midrule
% \multicolumn{3}{l}{\emph{Visual input - Resolution (Vision2Code, gpt-5.3-chat-latest)}} \\
% $256 \times 256$ (default)           & 0.305   & ref \\
% $512 \times 512$                            & 0.296   & -0.009 \\
% $768 \times  768$                           & 0.341   & 0.036 \\
% \midrule
% \multicolumn{3}{l}{\emph{Visual input - Resolution (Vision2Code, gpt-5.3-thinking)}} \\
% $256 \times 256$ (default)           & 0.330   & ref \\
% $512 \times 512$                            & 0.333   & 0.003 \\
% $768 \times  768$                           & 0.275   & -0.055 \\
\bottomrule
\end{tabular}
\end{table}

Table~\ref{tab:ablation} reports two scoring-protocol ablations on the Vision2Code task. (i) The \emph{24-axial vs.\ single-axis IoU} contrast quantifies how much credit is recovered by accounting for wrong-plane outputs (\S\ref{sec:exp_failure_analysis}, $L_1$ spatial-anchoring failure mode): when the gap is large, the model's geometry is approximately correct but emitted on a non-canonical workplane. (ii) The \emph{single-view vs.\ multi-view} contrast quantifies how much of a model's score is driven by the four-view evidence specifically, isolating the cost of removing back-side and depth cues from the perception input.

  Table~\ref{tab:ablation_img_size} reports an image-resolution ablation on the Vision2Code task, split by reasoning mode. (i) For the no-thinking model, increasing
  input resolution from 256 to 768 px lifts total score from 0.305 to 0.341 (best at 768 px), with IoU rising from 0.117 to a peak of 0.150 at 512 px (0.142 at 768
  px): when more pixels are available, the VLM resolves geometry that is sub-pixel at coarse scales and emits it as correct code. (ii) For the thinking model, the
  same increase is non-monotonic — IoU and total peak at the mid-resolution (512 px: 0.180 and 0.333) and then collapse at 768 px (IoU 0.117, total 0.275) — isolating
   an over-elaboration regime in which high-resolution visual detail amplifies into globally incorrect reasoning chains.

\begin{table}[t]
    \centering
    \small
    \setlength{\tabcolsep}{6pt}
    \renewcommand{\arraystretch}{1.15}
    \caption{\textbf{Visual resolution ablations.} Vision2Code task score on BenchCAD across different input image resolutions. Best per block in \textbf{bold}.}
    \label{tab:ablation_img_size}
    \begin{tabular}{l c c c c c c}
    \toprule
    Model & size (px) & IoU$\uparrow$ & CD$\downarrow$ & op\_F1$\uparrow$ & ess\_score$\uparrow$ & total$\uparrow$ \\
    \midrule
    \multirow{3}{*}{gpt-5.3-chat}
      & 256 & 0.117          & 0.123          & 0.494          & 0.73          & 0.305 \\
      & 512 & \textbf{0.150} & 0.181          & \textbf{0.499} & 0.63          & 0.296 \\
      & 768 & 0.142          & \textbf{0.115} & 0.492          & \textbf{0.82} & \textbf{0.341} \\
    \midrule
    \multirow{3}{*}{gpt-5.3-thinking}
      & 256 & 0.166          & 0.158          & 0.496          & \textbf{0.76} & 0.330 \\
      & 512 & \textbf{0.180} & 0.186          & \textbf{0.500} & 0.73          & \textbf{0.333} \\
      & 768 & 0.117          & \textbf{0.152} & 0.464          & 0.67          & 0.275 \\
    \bottomrule
    \end{tabular}
  \end{table}

\section{CadQuery Operation Coverage}
\label{app:ops}
\begin{table}[t]
\centering
\small
\setlength{\tabcolsep}{5pt}
\renewcommand{\arraystretch}{1.05}
\caption{\textbf{Per-category CadQuery operation coverage.} Distinct CadQuery method invocations per category in BenchCAD vs.\ the closest prior CadQuery corpus (CADEvolve), measured by the unified static-analysis rule (App.~\ref{app:ops}). CADEvolve releases only $46$ hand-written seed programs as source code (the expanded $\sim$$1.3$M corpus is shipped only as embeddings); its column is therefore a lower bound. Saturation curves and full op lists for all corpora (CAD-Recode, cadrille, GenCAD-Code, CADPrompt) are in App.~\ref{app:ops}.}
\label{tab:op_coverage}
\begin{tabular}{@{}lcc@{\hskip 1.4em}l@{}}
\toprule
Category & BenchCAD & CADEvolve & Examples (BenchCAD) \\
\midrule
Primitives           & 4   & 1   & box, cylinder, sphere, makeTorus \\
2D sketch profile    & 15  & 13  & circle, ellipse, spline, slot2D, threePointArc \\
Solid-forming        & 6   & 3   & extrude, revolve, sweep, loft, twistExtrude \\
Advanced path        & 3   & 0   & makeHelix, makeLine, assembleEdges \\
Boolean              & 5   & 3   & cut, cutBlind, cutThruAll, union, intersect \\
Hole features        & 3   & 3   & hole, cboreHole, cskHole \\
Finishing            & 2   & 1   & fillet, chamfer \\
Arrays / Mirrors     & 4   & 3   & polarArray, rarray, mirrorX, mirrorY \\
Transforms           & 3   & 3   & transformed, rotate, workplane \\
Sketch builder API   & 1   & 0   & placeSketch \\
\midrule
\textbf{Total}       & \textbf{46} & \textbf{30} & \\
\bottomrule
\end{tabular}
\end{table}

Table~\ref{tab:op_coverage} reports the per-category breakdown.

\paragraph{Static analysis protocol.}
For each released CadQuery corpus we extract distinct method invocations using the regex \texttt{$\backslash$.([A-Za-z\_]\textbackslash w*)(} on the \texttt{.py} text. We then EXCLUDE three groups: (a) class/type names (\texttt{Workplane}, \texttt{Vector}, \texttt{Wire}, \texttt{Solid}, \texttt{Sketch}, \texttt{Compound}, \texttt{Edge}, \texttt{Face}, \texttt{Plane}, \texttt{Location}, \texttt{Shape}, \texttt{Edges}, \texttt{Vertices}); (b) selectors / accessors that do not modify geometry (\texttt{face}, \texttt{faces}, \texttt{edges}, \texttt{vertices}, \texttt{wires}, \texttt{val}, \texttt{vals}, \texttt{first}, \texttt{last}, \texttt{tag}, \texttt{newObject}, \texttt{copyWorkplane}, \texttt{plane}); and (c) Python stdlib / numpy / utility helpers (\texttt{append}, \texttt{join}, \texttt{format}, \texttt{sin}, \texttt{cos}, \texttt{multiply}, \ldots). The same EXCLUDE list is applied uniformly across BenchCAD, CADEvolve seeds, CAD-Recode/cadrille, GenCAD-Code, and CADPrompt. Saturation curves: CAD-Recode at $1{,}274$ \texttt{.py} (op count constant from $200$ files onward); GenCAD-Code at $2{,}000$ streamed samples (constant from $100$ samples onward).

\paragraph{Tier B (sketch+extrude IR works).}
DeepCAD, Fusion360 Gallery, and Text2CAD release a tokenized intermediate representation rather than executable CadQuery, so the CadQuery static-analysis rule does not apply directly. We instead report the IR's primitive-token alphabet: DeepCAD encodes parts as \{\texttt{L}, \texttt{A}, \texttt{C}, \texttt{EXT}\} plus two control tokens \{\texttt{SOL}, \texttt{EOS}\} (\citet{Wu2021}, Sec.~3); Fusion360 Gallery r1.0.1 specifies $8$ sketch curve types (\texttt{SketchLine}, \texttt{SketchArc}, \texttt{SketchCircle}, \texttt{ConicCurve}, \texttt{Ellipse}, \texttt{EllipticalArc}, \texttt{FittedSpline}, \texttt{FixedSpline}) plus extrude with $4$ boolean variants \citep{Willis2021fusion}; Text2CAD reuses the CAD-SIGNet tokenizer ($\sim$$17$-token alphabet over end-of-X markers, quantized coords, and $4$ extrude booleans). These counts are not directly comparable to method counts on CadQuery \texttt{.py} releases; in Table~\ref{tab:cmp_prior} we accordingly summarize each work as \emph{narrow} (sketch+extrude IR or a small CadQuery subset) or \emph{broad} (wide CadQuery API).

\paragraph{CADEvolve caveat.}
CADEvolve's Hugging Face release (\texttt{kulibinai/cadevolve}, $\sim$$2$M rows, $4.7$\,GB) contains only \texttt{.npy} sentence embeddings keyed by component name, not Python source. The full $\sim$$1.3$M expanded scripts are referenced in the paper but are not part of the public release. We therefore report CADEvolve's operation count from the $46$ hand-written seed programs in \texttt{evolution/cadquery\_examples.txt} of the GitHub repo \texttt{zhemdi/CADEvolve}, yielding $30$ distinct ops --- a strict lower bound on the full evolved corpus.

% Case 167 — motor_end_cap_bore_widen failure analysis (paper-section form).
% Drop-in: \input{...} from main paper. Requires (already loaded by main):
%   listings, xcolor, tabularx, booktabs, caption, graphicx, tcolorbox.
% No \documentclass / \begin{document} here.

% --- code-block style (define once; safe if redefined) --------------------
\providecommand{\bad}[1]{\colorbox{badbg}{\color{annot}\bfseries\ttfamily #1}}
\providecommand{\good}[1]{\colorbox{goodbg}{\color{annotgood}\bfseries\ttfamily #1}}
\providecommand{\arrowL}[1]{{\color{annot}\(\ \leftarrow\ \)\textit{#1}}}
\providecommand{\arrowG}[1]{{\color{annotgood}\(\ \leftarrow\ \)\textit{#1}}}
\definecolor{cqkw}{HTML}{1F7A3A}
\definecolor{cqstr}{HTML}{B85A5A}
\definecolor{cqcom}{HTML}{888888}
\definecolor{badbg}{HTML}{FFE5E5}
\definecolor{goodbg}{HTML}{E5F4E0}
\definecolor{annot}{HTML}{C0392B}
\definecolor{annotgood}{HTML}{1F7A3A}
\lstdefinelanguage{cqpython}[]{Python}{%
  morekeywords={cq,Workplane,faces,workplane,cylinder,hole,circle,extrude,%
    edges,chamfer,fillet,polarArray,cut,union,result,show\_object},%
}
\lstset{%
  language=cqpython,%
  basicstyle=\ttfamily\fontsize{7.6}{9.0}\selectfont,%
  keywordstyle=\color{cqkw}\bfseries,%
  stringstyle=\color{cqstr},%
  commentstyle=\color{cqcom}\itshape,%
  showstringspaces=false,%
  breaklines=true,%
  xleftmargin=0pt,%
  framexleftmargin=0pt,%
  frame=none,%
  backgroundcolor=\color{gray!4},%
  escapeinside={(*@}{@*)},%
}

\section{Case Study: \texttt{motor\_end\_cap\_bore\_widen} (Case 167)}
\label{app:case_167}

\paragraph{Setup.}
Instruction: \emph{``Add a central cylindrical through-cut with a $20.00$\,mm radius through the flange.''}
Baseline IoU(orig,\,GT) $= 0.941$. Three of the four 2026 frontier models we tested return \emph{identical} STEP outputs at IoU(gen,\,GT) $= 0.961$ --- the cut takes effect on the base flange but \emph{not} on the upper boss, because the cut operation is placed in the middle of the build chain instead of as a final \texttt{result.cut(\dots)}.

\begin{figure}[h]
  \centering
  \begin{minipage}[t]{0.45\linewidth}
    \centering
    \includegraphics[width=\linewidth]{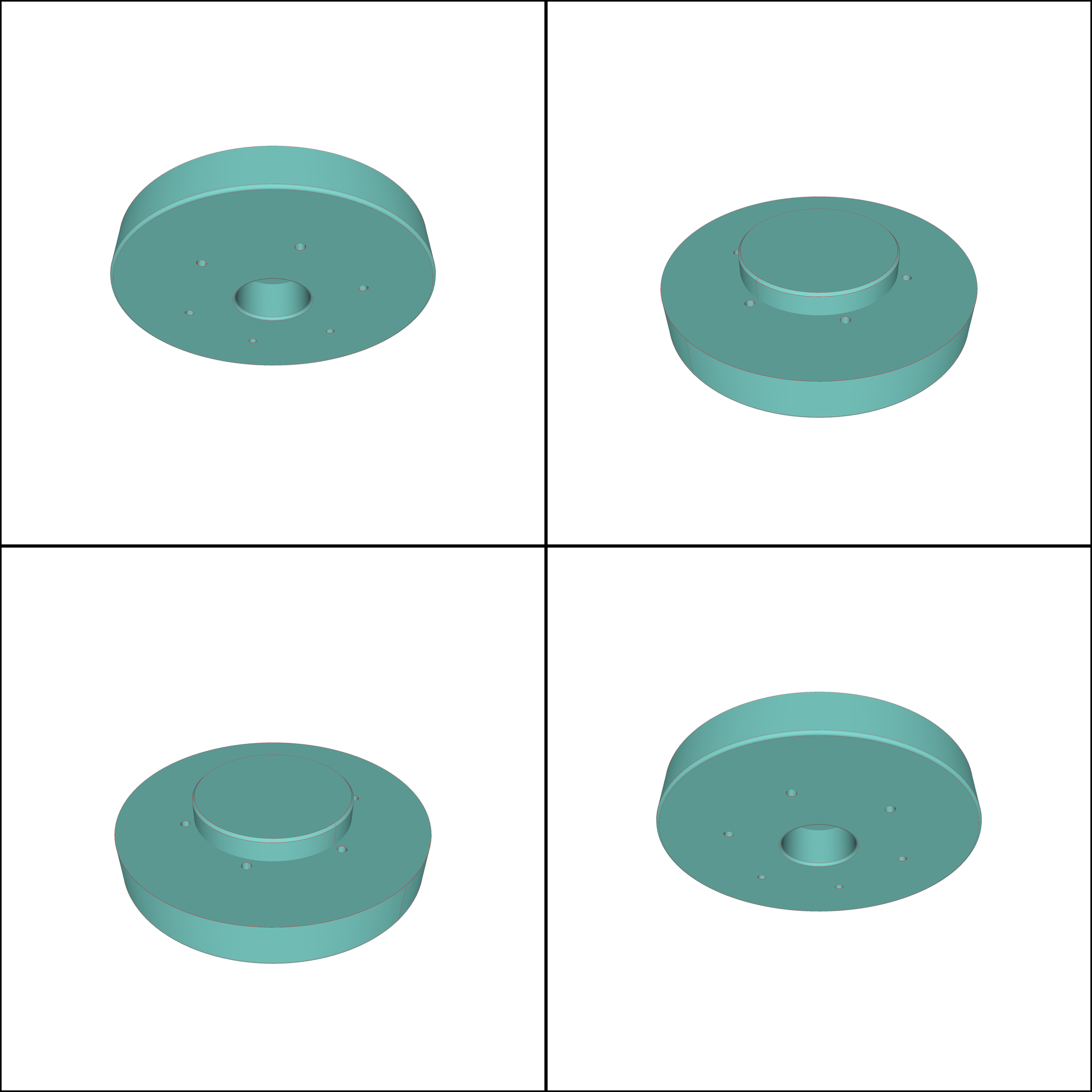}
    \caption*{\textbf{Original.} Motor end cap with $32.9$\,mm shaft hole through the flange and a $35.6$\,mm-radius solid boss on top.}
  \end{minipage}\hfill
  \begin{minipage}[t]{0.45\linewidth}
    \centering
    \includegraphics[width=\linewidth]{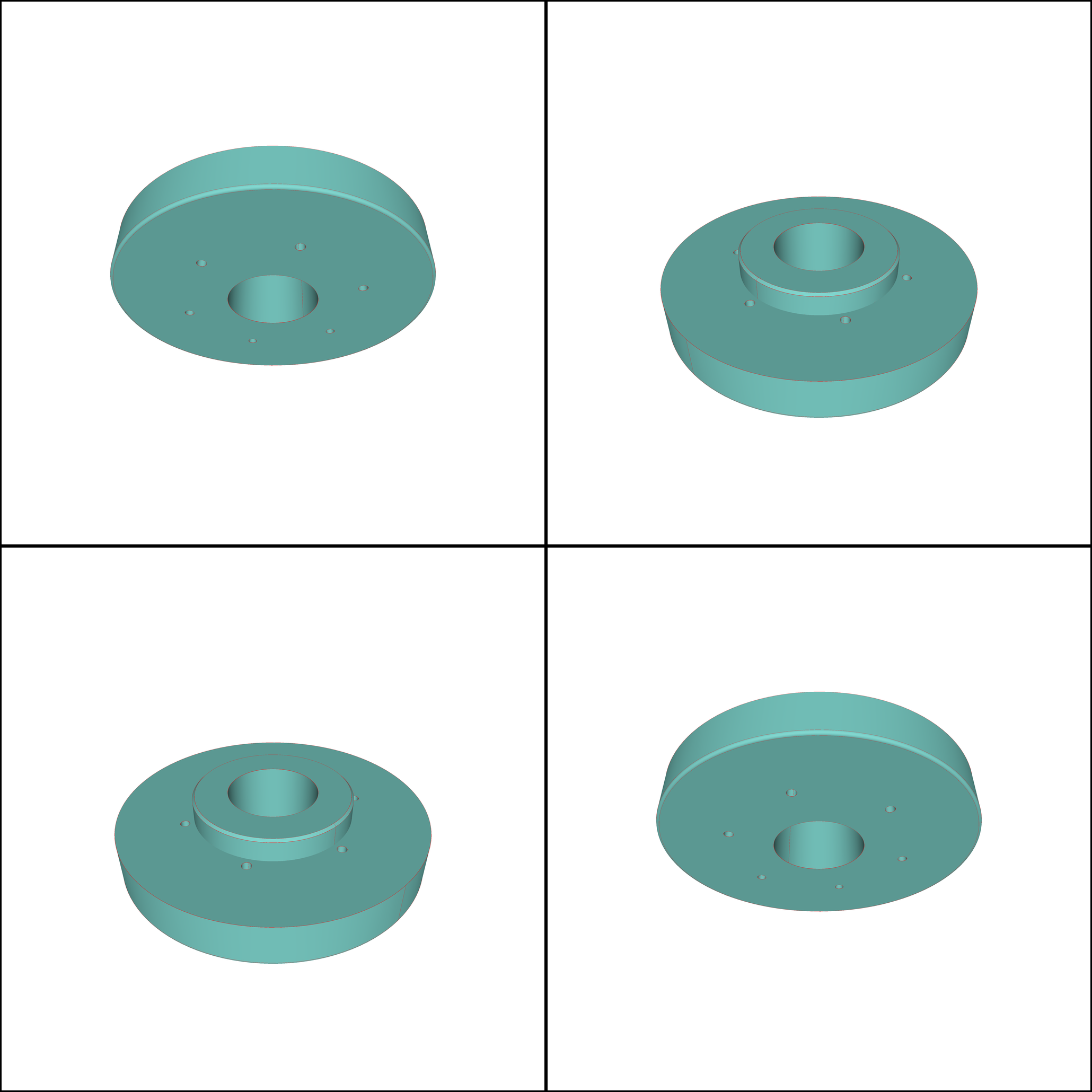}
    \caption*{\textbf{Ground truth.} Same part with the central $20$\,mm-radius cylindrical bore extended through both the flange and the boss.}
  \end{minipage}
  \caption{\textbf{Case 167: original vs.\ ground-truth target.} The intended edit pierces every layer of the build chain.}
  \label{fig:case_167_geom}
\end{figure}

\begin{tcolorbox}[
    title={\texttt{Ground truth (IoU $=$ 1.0)} },
    colback=gray!4,
    colframe=gray!45,
    coltitle=black,
    colbacktitle=gray!4,
    fonttitle=\bfseries,
    breakable,
    sharp corners,
    boxrule=0.5pt,
]
\begin{lstlisting}
result = (
    cq.Workplane("XY")
    .cylinder(19.9, 69.6)            # base flange
    .faces(">Z").workplane()
    .hole(32.9)                      # original shaft hole
    .faces(">Z").workplane()
    .circle(35.6).extrude(10.4)      # upper boss (sits on flange)
    .edges(">Z").chamfer(1.1)
    .edges("<Z").fillet(1.3)
    .faces(">Z").workplane()
    .polarArray(39.79, 0, 360, 6)
    .hole(4.2)                       # 6 vent holes
)
# --- the edit: a final cylinder cut on the assembled solid
(*@\good{result = result.cut(cq.Workplane('XY').cylinder(100.0, 20.0))}@*)
show_object(result)
\end{lstlisting}
\end{tcolorbox}
\textit{Idea.} Build the whole part first, then \texttt{.cut} a $r{=}20$ cylinder of height $100$ \emph{after} the build chain closes. The oversized height ($100\!\gg\!30$) guarantees the cut pierces both the $19.9$\,mm flange and the $10.4$\,mm boss in one call.

\begin{tcolorbox}[
    title={\texttt{GPT-5.3 (IoU $=$ 0.961)}},
    colback=gray!4,
    colframe=gray!45,
    coltitle=black,
    colbacktitle=gray!4,
    fonttitle=\bfseries,
    breakable,
    sharp corners,
    boxrule=0.5pt,
]
\begin{lstlisting}
result = (
    cq.Workplane("XY")
    .cylinder(19.9, 69.6)
    .faces(">Z").workplane()
    .hole(32.9)
    .faces(">Z").workplane()
    (*@\bad{.hole(40.0)}@*)                      (*@\arrowL{cuts only the flange}@*)
    .faces(">Z").workplane()
    .circle(35.6).extrude(10.4)      (*@\arrowL{boss now seals the cut}@*)
    .edges(">Z").chamfer(1.1)
    .edges("<Z").fillet(1.3)
    .faces(">Z").workplane()
    .polarArray(39.79, 0, 360, 6)
    .hole(4.2)
)
show_object(result)
\end{lstlisting}
\end{tcolorbox}
\textit{Bug.} \texttt{.hole(40.0)} is inserted between the shaft hole and the boss \texttt{union}. It cuts a $40$\,mm-diameter through-hole on the flange's top face only; CadQuery's \texttt{.hole()} does \emph{not} retroactively pierce a primitive \texttt{.union}-ed in afterwards. The boss is solid in its central $40$\,mm region.

\begin{tcolorbox}[
    title={\texttt{GPT-5.3 thinking (IoU $=$ 0.961)}},
    colback=gray!4,
    colframe=gray!45,
    coltitle=black,
    colbacktitle=gray!4,
    fonttitle=\bfseries,
    breakable,
    sharp corners,
    boxrule=0.5pt,
]
\begin{lstlisting}
result = (
    cq.Workplane("XY")
    .cylinder(19.9, 69.6)
    .faces(">Z").workplane()
    (*@\bad{.hole(40.0)}@*)                      (*@\arrowL{rewrote orig 32.9 to 40}@*)
    .faces(">Z").workplane()
    (*@\bad{.hole(32.9)}@*)                      (*@\arrowL{redundant: 32.9 sub of 40}@*)
    .faces(">Z").workplane()
    .circle(35.6).extrude(10.4)      (*@\arrowL{boss seals it again}@*)
    ...
)
\end{lstlisting}
\end{tcolorbox}
\textit{Bug.} Two violations of the minimal-edit rule \emph{plus} the same boss-seal problem: (i) modifies the existing \texttt{.hole(32.9)} (should stay) and (ii) re-adds a redundant \texttt{.hole(32.9)} that is fully contained inside the new $40$ hole, hence a no-op. The geometric outcome is identical to GPT-5.3 above.

\begin{tcolorbox}[
    title={\texttt{Gemini 3.1} (IoU $=$ 0.961)},
    colback=gray!4,
    colframe=gray!45,
    coltitle=black,
    colbacktitle=gray!4,
    fonttitle=\bfseries,
    breakable,
    sharp corners,
    boxrule=0.5pt,
]
\begin{lstlisting}
result = (
    cq.Workplane("XY")
    .cylinder(19.9, 69.6)
    .faces(">Z").workplane()
    (*@\bad{.hole(40.0)}@*)                      (*@\arrowL{replaced orig hole, in-place}@*)
    .faces(">Z").workplane()
    .circle(35.6).extrude(10.4)      (*@\arrowL{same boss-seal failure}@*)
    .edges(">Z").chamfer(1.1)
    .edges("<Z").fillet(1.3)
    .faces(">Z").workplane()
    .polarArray(39.79, 0, 360, 6)
    .hole(4.2)
)
show_object(result)
\end{lstlisting}
\end{tcolorbox}
\textit{Bug.} The most concise of the three wrong rewrites: directly bumps the existing shaft hole from $32.9$ to $40$. Same structural error: cut applied before the boss is added, so the boss reseals the central column.

\paragraph{Failure pattern (F08, L4: spatial reasoning $+$ code synthesis).}
All three $\mathrm{IoU}\!=\!0.961$ models produce the \emph{identical} STEP file: the bore is widened to $40$\,mm in the $19.9$\,mm flange but the $10.4$\,mm-tall boss above it remains a solid plug. They mistake \texttt{.hole(d)} (a workplane-local cut on the current solid) for a global through-cut. The correct CadQuery idiom is to apply a final \texttt{result.cut(cq.Workplane(\dots).cylinder(big\_h, r))}, which pierces every layer regardless of when each layer joined the build chain.

\section{Reproduction}
BenchCAD provides three independently runnable benchmarks:
\texttt{CodeEdit/}, \texttt{CodeGen/}, and \texttt{CodeQA/}, all under a unified
Python~3.11 environment managed by \texttt{uv}. After running \texttt{uv sync} and
adding the required model API keys to \texttt{.env}, the full benchmark can be
reproduced with
\texttt{uv run python run\_all.py --config prod}. Each task can also be executed
individually with its own \texttt{main.py} and configuration file. Benchmark data
is downloaded from the Hugging Face dataset \texttt{BenchCAD/BenchCAD}, and each
task includes a small \texttt{test\_data/} split for smoke testing. Model names,
decoding options, and output directories are specified in YAML configuration
files. We use deterministic decoding where supported and report the same task
metrics as in the main paper: normalized voxel IoU for CodeEdit, voxel IoU for
CodeGen, and mean symmetric ratio accuracy for CodeQA.

\clearpage
\section{Ethical Considerations and Broader Impact}
\label{app:ethics}

BenchCAD contains procedurally generated parametric CAD parts and references public engineering standards by code and name.
The dataset contains no personal data, no proprietary designs, and no safety-sensitive specifications.
Potential risks of CAD-capable models more broadly --- e.g.\ accelerated reverse-engineering of regulated components --- are not specific to BenchCAD and are best addressed at the model-deployment layer rather than the benchmark layer.
We see no negative societal impact specific to this work that is not already present in the underlying open-source CAD ecosystem.

% add if submit to nips
% \newpage
% \input{checklist}

\end{document}